\documentclass[5p,times]{elsarticle}

\usepackage{amsmath}
\usepackage{amssymb}
\usepackage{amsfonts}
\usepackage{textcomp}

\usepackage{algorithm}
\usepackage{algpseudocode}

\usepackage{multirow}
\usepackage{longtable}
\usepackage{array}
\usepackage{makecell}
\usepackage{tabularx}

\newcolumntype{P}[1]{>{\centering\arraybackslash}p{#1}}

\usepackage{caption}
\usepackage{subcaption}
\usepackage{multirow}
\usepackage{booktabs}
\usepackage[table]{xcolor}

\definecolor{stdred}{RGB}{245,210,210}     
\definecolor{causalred}{RGB}{235,170,170}  
\usepackage[final]{microtype}   
\usepackage[english]{babel}     
\usepackage{xurl}               
\usepackage{xcolor}
\usepackage[colorlinks=true,allcolors=blue]{hyperref}

\usepackage{pifont}
\usepackage[T1]{fontenc}
\usepackage{textcomp}
\usepackage[switch]{lineno}

\usepackage{listings}
\usepackage[most]{tcolorbox}

\definecolor{promptheader}{RGB}{0,0,0}          
\definecolor{promptback}{RGB}{250,250,250}      
\definecolor{responseheader}{RGB}{70,130,180}   
\definecolor{responseback}{RGB}{235,245,255}    

\newtcolorbox{promptbox}[1]{
  enhanced,
  colback=promptback,
  colframe=promptheader,
  coltitle=white,
  fonttitle=\bfseries,
  title={#1},
  arc=2pt,
  boxrule=1pt,
  left=5pt, right=5pt, top=5pt, bottom=5pt,
  sharp corners=south, 
  drop fuzzy shadow
}

\newtcolorbox{responsebox}[1]{
  enhanced,
  colback=responseback,
  colframe=responseheader,
  coltitle=white,
  fonttitle=\bfseries,
  title={#1},
  arc=2pt,
  boxrule=1pt,
  left=5pt, right=5pt, top=5pt, bottom=5pt,
  drop fuzzy shadow
}

\lstdefinelanguage{json}{
    basicstyle=\small\ttfamily,
    columns=fullflexible,
    showstringspaces=false,
    commentstyle=\color{gray},
    keywordstyle=\color{blue},
    stringstyle=\color{red!70!black},
    breaklines=true,
    frame=none,
}

\begin{document}

\let\WriteBookmarks\relax
\def\floatpagepagefraction{1}
\def\textpagefraction{.001}



\title{ORACAL: A Robust and Explainable Multimodal Framework for Smart Contract Vulnerability Detection with Causal Graph Enrichment} 




%

\affiliation[1]{organization={Information Security Lab, University of Information Technology},
    city={Ho Chi Minh City},
    country={Vietnam}}
\affiliation[2]{organization={Vietnam National University},
    city={Ho Chi Minh City},
    country={Vietnam}}
\affiliation[3]{organization={School of Computer Science and Information Technology, Adelaide University},
    city={Adelaide},
    country={Australia}}

\author[1,2]{Tran Duong Minh Dai}
\ead{22520183@gm.uit.edu.vn}

\author[3]{Triet Huynh Minh Le}
\ead{triet.h.le@adelaide.edu.au}

\author[3]{M. Ali Babar}
\ead{ali.babar@adelaide.edu.au}

\author[1,2]{Van-Hau Pham}
\ead{haupv@uit.edu.vn}
\author[1,2]{Phan The Duy\corref{cor1}}
\ead{duypt@uit.edu.vn}
\cortext[cor1]{Corresponding author}

\begin{abstract}

Smart contract vulnerabilities arise from complex interactions among control flow, data dependencies, and inter-contract calls. Although Graph Neural Networks (GNNs) have shown promise for vulnerability detection, homogeneous graph models fail to capture multi-relational dependencies, heterogeneous approaches lack deep semantic understanding, and most black-box models provide no explainable evidence, hindering trust in professional audits. This paper aims to overcome the limitations of existing GNN-based detectors by (i) capturing heterogeneous structural semantics across multiple graph modalities, (ii) distinguishing genuine vulnerability indicators from spurious correlations under adversarial conditions, and (iii) providing interpretable, subgraph-level explanations for detected vulnerabilities. We propose ORACAL (Observable RAG-enhanced Analysis with CausAL reasoning), a heterogeneous multimodal graph learning framework that jointly integrates Control Flow Graph (CFG), Data Flow Graph (DFG), and Call Graph (CG). ORACAL selectively enriches critical subgraphs with expert-level security context via Retrieval-Augmented Generation (RAG) and Large Language Models (LLMs), and employs a causal attention mechanism to disentangle true vulnerability signals from confounding factors. Transparency is achieved through PGExplainer, which generates subgraph-level explanations identifying vulnerability-triggering paths. Experiments on large-scale benchmarks show that ORACAL achieves state-of-the-art performance, surpassing MANDO-HGT, MTVHunter, GNN-SC, and SCVHunter by up to 39.6 percentage points, with a peak Macro F1 of 91.28\%. Strong out-of-distribution generalization is demonstrated with 91.8\% on CGT Weakness and 77.1\% on DAppScan. PGExplainer attains 32.51\% Mean Intersection over Union (MIoU) against manually annotated paths. Under adversarial attack, ORACAL limits performance degradation to approximately 2.35\% F1 decrease with an Attack Success Rate of only 3\%, compared to 10.91\% to 18.73\% for competing baselines. In conclusion, ORACAL demonstrates that combining heterogeneous graph learning, causal reasoning, and RAG-driven semantic enrichment yields a robust, explainable, and adversarially resilient framework for smart contract vulnerability detection.

\end{abstract}




\maketitle

\section{Introduction}
\label{sec:introduction}

Smart contracts have become a foundational element of the modern blockchain ecosystem, driving widespread adoption in Decentralized Finance (DeFi) and Non-Fungible Tokens (NFTs). Their ability to automatically execute business logic without trusted intermediaries has fostered innovation; however, their immutable nature presents a \textquotedbl{}double-edged sword\textquotedbl{}. Once deployed, code cannot be easily patched, meaning vulnerabilities can lead to permanent and catastrophic financial losses.

In recent years, high-profile incidents such as the Cetus Protocol exploit, where an integer overflow allowed attackers to drain over \$220 million \cite{securityweek_cetus}, the Zoth Protocol breaches involving combined logic and governance failures \cite{halborn_zoth}, and the TrueBit protocol's mathematical overflow flaw resulting in \$26.6 million in losses \cite{coindesk2026truebit}, have demonstrated that vulnerabilities are evolving beyond simple logic errors toward complex economic state manipulations. DeFi liquidity protocols Aperture Finance and Swapnet were similarly exploited due to insufficient input validation and arbitrary external call vulnerabilities in early 2026 \cite{blocksec2026swapnet}. Reports indicate that DeFi exploits caused over \$2.9 billion in losses in 2025 alone \cite{slowmist2025}, highlighting an urgent need for robust automated security analysis in smart contract systems.

Traditional vulnerability detection methods, such as static analysis tools like Slither or Mythril, rely heavily on predefined rules and pattern matching. While effective for known vulnerability classes, they struggle with complex, novel logic errors and lack deep semantic understanding \cite{hejazi2025comprehensive,iuliano2026smart}. Deep learning approaches have emerged as a promising alternative, particularly in identifying vulnerabilities through graph representations of code \cite{bresil2025deep,crisostomo2025machine}. 

However, homogeneous graphs capture only limited aspects of the code by modeling all nodes and relations uniformly, overlooking the diverse semantic roles of program constructs \cite{wang2025review}. Recent advances utilizing heterogeneous graphs \cite{luo2024scvhunter,meng2025smartscope} have shown improved performance but often still treat code semantics superficially. More critically, heterogeneous graph neural networks are highly susceptible to adversarial attacks \cite{zhao2024hgattack,wang2025heta,li2025metapath}, where minor structural perturbations can mislead the model due to its reliance on shallow structural features without robust grounding in the code's actual semantic intent and security context.

The rise of Large Language Models (LLMs) offers a new frontier, as models like GPT-4 or CodeBERT can capture intricate code semantics. However, LLM-based vulnerability detectors do not inherently resolve the adversarial robustness issue: they can be misled by carefully crafted inputs or semantically equivalent but security-critical code transformations \cite{yang2024assessing,jabbar2025red}. Moreover, when used in isolation, they suffer from lack of domain-specific grounding leading to hallucinations \cite{li2025scalm}, and a black-box nature that fails to provide explainable evidence for predictions \cite{al2025emerging}. In security audits, identifying a bug is not enough; auditors need to know \textit{why} it is a bug and \textit{where} the root cause lies \cite{ghanmi2025evaluating}.

To address these gaps, we propose \textbf{ORACAL} (Observable RAG-enhanced Analysis with Causal Reasoning), a novel framework for trustworthy smart contract vulnerability detection. ORACAL combines heterogeneous graph representations with retrieval-augmented semantic grounding. Specifically, it enriches critical nodes with structured semantic context generated by an LLM conditioned on a curated knowledge base of Solidity vulnerabilities, promoting causally relevant feature attribution over spurious correlations.

Furthermore, we introduce a causal attention mechanism to mitigate the \textquotedbl{}Clever Hans\textquotedbl{} effect \cite{lapuschkin2019unmasking}, in which models may rely on superficial artifacts, such as node counts, common control flow graph patterns, or frequent opcodes, rather than actual security logic. This design ensures that the model focuses on causal vulnerability patterns instead of spurious correlations in the dataset, thereby strengthening robustness against adversarial perturbations. To enhance interpretability, ORACAL adopts PGExplainer \cite{luo2020parameterized} to generate subgraph-level explanations that identify the most influential nodes and edges contributing to each vulnerability prediction, enabling auditors to quickly verify the root cause of detected issues.

This paper makes the following key contributions:
\begin{itemize}
    \item \textbf{Heterogeneous Multimodal Graph Framework:} We propose a novel method to construct a heterogeneous graph (combining CFG, DFG, and Call Graph) and enrich its critical subgraphs using an LLM-based RAG pipeline. This injects expert-level security knowledge directly into the graph structure.
    \item \textbf{Causal Attention Learning:} We design a dual-branch graph neural network that explicitly disentangles causal features (derived from RAG enrichment) from spurious features (node contents), significantly improving generalizability across different datasets.
    \item \textbf{Explainability and Trustworthiness:} We employ PGExplainer for subgraph-level explanations, achieving 32.51\% MIoU against manually annotated vulnerability triggering paths. Adversarial robustness evaluation shows that ORACAL reduces the Attack Success Rate (ASR) to approximately 3\%, compared to up to 19\% for existing detectors.
    \item \textbf{Comprehensive Evaluation:} We evaluate ORACAL on two in-domain datasets (SoliAudit \cite{liao2019soliaudit}, CGTWeakness \cite{di2023consolidation}) and two out-of-distribution benchmarks (DAppScan \cite{zheng2024dappscan}, LLMAV \cite{salzano2025empirical}). ORACAL achieves a Macro F1 of 91.28\% on the primary test set, improving over prior detectors by up to 39 percentage points, and maintains strong generalization on OOD datasets (91.8\% on CGTWeakness, 77.1\% on DAppScan).
\end{itemize}

\textbf{Paper structure}. Section \ref{sec:related_work} reviews existing approaches to smart contract vulnerability detection and positions our work relative to prior methods. Section \ref{sec:methodology} presents the ORACAL framework in detail. Section \ref{sec:experiment} describes our experimental design and evaluation results. Section \ref{sec:threats_validity} analyzes threats to the validity of our study, and Section \ref{sec:conclusion} concludes the paper and outlines future research directions.

\section{Related Work}
\label{sec:related_work}

\begin{table*}[t]
  \centering
  \caption{Comparison of Our Work with Related Smart Contract Vulnerability Detection Methods}
  \label{tab:methods-comparison}
  \resizebox{\textwidth}{!}{%
    \begin{tabular}{lclllc}
      \hline
      \textbf{Method} & \textbf{Year} & \textbf{Category} & \textbf{Technique} & \textbf{Semantic} & \textbf{Explainability} \\
      \hline
      Oyente~\cite{luu2016making}     & 2016 & Static Analysis & Symbolic Execution & No & Yes \\
      \hline
      Slither~\cite{feist2019slither}    & 2018 & Static Analysis & Rule-based & No & Yes \\
      \hline
      Peculiar~\cite{wu2021peculiar}   & 2019 & Deep Learning & Homogeneous Graph & No & No \\
      \hline
      GNN-SC~\cite{cheong2024gnn}     & 2021 & Deep Learning & Homogeneous Graph & No & No \\
      \hline
      MANDO-HGT~\cite{nguyen2023mando}  & 2023 & Deep Learning & Heterogeneous Graph & No & No \\
      \hline
      SCVHunter~\cite{luo2024scvhunter}  & 2024 & Deep Learning & Heterogeneous Graph & No & No \\
      \hline
      MTVHunter~\cite{sun2025mtvhunter}  & 2025 & Deep Learning & Heterogeneous Graph & Static Knowledge & No \\
      \hline
      \textbf{ORACAL (Ours)} & \textbf{2026} & \textbf{Deep Learning} & \textbf{Heterogeneous Multimodal Graph} & \textbf{Dynamic Knowledge} & \textbf{Yes} \\
      \hline
    \end{tabular}
  }
\end{table*}

This section reviews prior work on smart contract vulnerability detection. As summarized in Table~\ref{tab:methods-comparison}, we categorize existing approaches into three groups: Rule-based Static Analysis, Homogeneous Graph Learning, and Heterogeneous Graph Learning. Compared to these methods, our proposed ORACAL framework introduces a Heterogeneous Multimodal Graph enriched with dynamic RAG-based security knowledge and provides explicit explainability via causal attention and subgraph-level explanations, which are capabilities absent from all prior approaches.

\subsection{Static Analysis}

As summarized in Table~\ref{tab:methods-comparison}, static analysis tools exemplified by Slither \cite{feist2019slither} and Oyente \cite{luu2016making} analyze Solidity source code or EVM bytecode using symbolic execution, control flow reasoning, and predefined vulnerability patterns. These tools provide rule-level traceability and interpretable warnings, making them suitable for manual auditing workflows.

Nevertheless, their detection capability is inherently bounded by handcrafted rules and predefined signatures. As a result, they primarily identify known vulnerability patterns and struggle to generalize to unseen or evolving attack strategies. Ghaleb and Pattabiraman \cite{ghaleb2020effective} report substantial false-positive and false-negative rates across widely used smart contract analyzers. Their findings indicate that rule-based engines heavily rely on predefined vulnerability patterns, thereby limiting their reliability when analyzing complex real-world contracts.

\subsection{Graph Learning Approaches}

\textbf{Homogeneous Graph Models.} Early deep learning approaches model smart contracts as graphs for vulnerability detection. Peculiar \cite{wu2021peculiar} constructs a critical data flow graph emphasizing security-sensitive operations, while GNN-SC \cite{cheong2024gnn} applies Graph Convolutional Networks over control flow graphs for multi-label classification. Although these models improve generalization over static analyzers, they encode only a single type of structural dependency within a unified graph schema, limiting their ability to jointly reason across multiple semantic dimensions required for detecting complex vulnerabilities involving state transitions, cross-function interactions, or intertwined control and data dependencies.

\textbf{Heterogeneous Graph Models.}  
To overcome the structural constraints of homogeneous graphs, recent studies construct heterogeneous program graphs that integrate multiple semantic relations within a unified framework. MANDO-HGT \cite{nguyen2023mando} builds heterogeneous contract graphs (HCGs) from either source code or bytecode, incorporating control-flow and function-call relations, and employs heterogeneous graph transformers with customized meta-relations to model diverse node and edge types for both contract-level and line-level vulnerability detection. SCVHunter \cite{luo2024scvhunter} designs a heterogeneous semantic graph based on intermediate representations and applies a heterogeneous graph attention network to capture structural and semantic dependencies, while allowing lightweight expert guidance to emphasize critical nodes. These approaches demonstrate that multi-relational modeling provides richer contextual representations than single-structure graph encodings.

Despite these structural advances, heterogeneous graph models remain primarily syntax-driven. Node relationships are typically derived from static structural dependencies, without explicitly addressing noise interference or missing semantic information at the bytecode level. MTVHunter \cite{sun2025mtvhunter} introduces a multi-teacher framework to enhance bytecode vulnerability detection, incorporating an instruction denoising teacher and a semantic complementary teacher with neuron distillation to transfer opcode-level knowledge. However, its semantic enhancement remains largely static and task-specific, lacking mechanisms to dynamically adapt to evolving vulnerability patterns or higher-level reasoning requirements.

\subsection{Explainability and Robustness Challenges}

As highlighted in Table \ref{tab:methods-comparison}, a common failing across almost all graph-based deep learning methods is the lack of explainability. They operate as black boxes, providing a prediction without offering the \textquotedbl{}why,\textquotedbl{} which is a critical requirement for trust in security auditing. A comprehensive survey by Li et al. \cite{li2025can} highlights that existing explanation techniques for graph neural networks are still insufficient for delivering faithful and human-interpretable justifications. In the context of vulnerability detection, Chu et al. \cite{chu2024graph} propose a counterfactual explanation framework to identify minimal structural changes that alter model predictions, demonstrating that additional mechanisms are required to interpret graph-based detectors. Similarly, Cao et al. \cite{cao2024coca} introduce a causal explanation approach to improve both interpretability and robustness of GNN-based vulnerability detection systems. These studies indicate that explainability is not inherently supported by standard graph neural architectures and must be explicitly incorporated.

Robustness is another critical concern. Graph neural networks depend heavily on structural patterns learned from graph topology. Prior studies have shown that graph-based models can be sensitive to structural perturbations and spurious correlations \cite{xu2021robustness, wang2025heta,li2025metapath}. Minor modifications to nodes or edges may significantly alter predictions even when program semantics remain unchanged. This structural dependency makes heterogeneous graph models potentially vulnerable to adversarial manipulation, raising concerns for deployment in obfuscated or adversarial blockchain environments.

\subsection{Summary}
In summary, static analyzers are interpretable but rule-constrained, homogeneous graph models offer limited structural coverage, and heterogeneous models remain syntax-driven without intrinsic semantic reasoning. All face persistent challenges in explainability and robustness. ORACAL bridges this gap by integrating RAG-based LLM security knowledge into heterogeneous graph representations and employing causal attention to distinguish genuine vulnerability indicators from spurious correlations.

\section{Methodology}
\label{sec:methodology}

\begin{figure*}[t]
    \centering
    \includegraphics[width=\textwidth]{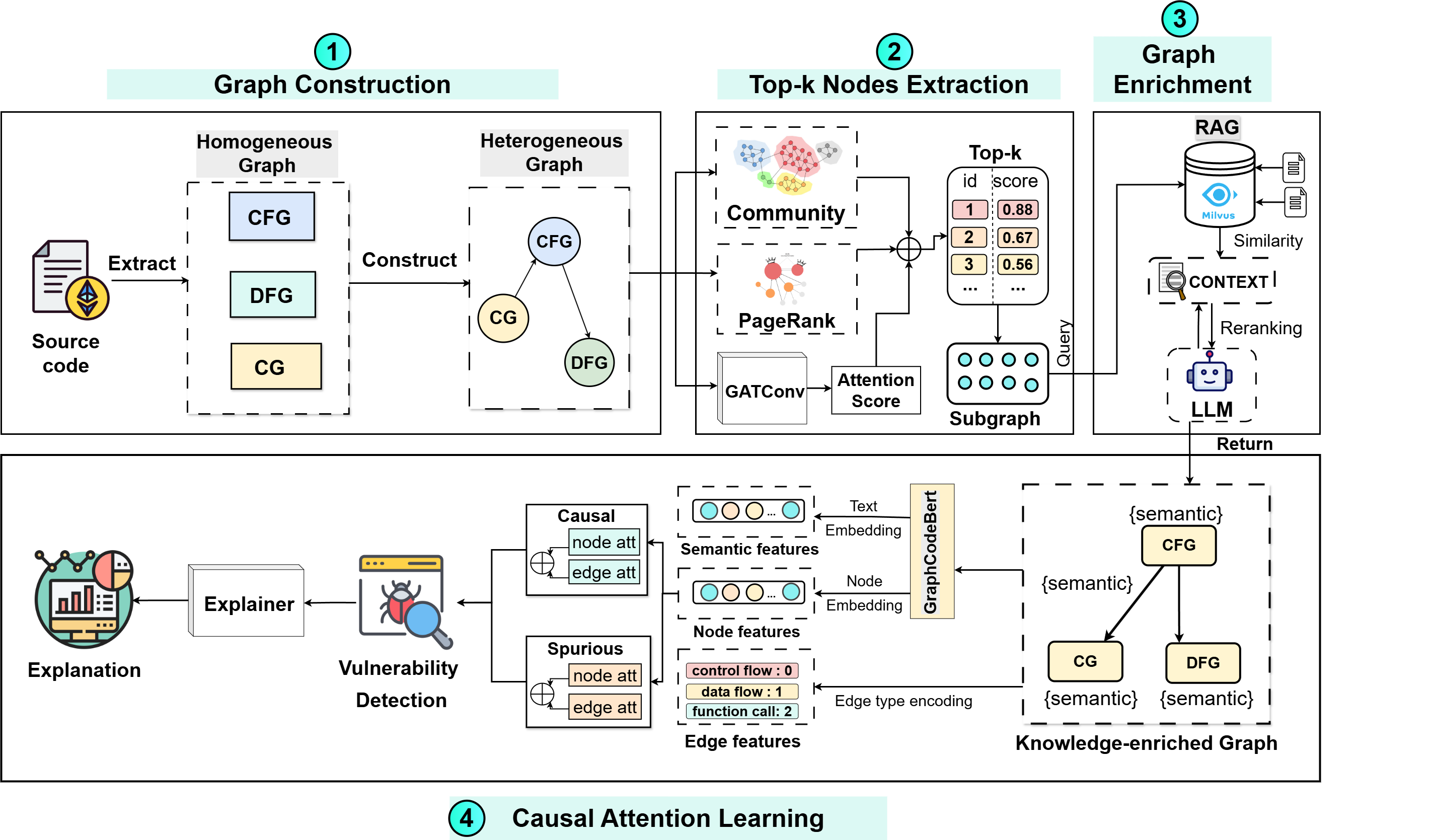}
    \caption{Overview of the ORACAL Framework. The pipeline proceeds from heterogeneous graph construction to identifying critical nodes, enriching them via RAG, and finally training a causal attention-based GNN.}
    \label{fig:model-overview}
\end{figure*}

We propose ORACAL (Observable RAG enhanced Analysis with CausAL reasoning), a novel framework that synergizes Heterogeneous Graph Neural Networks with Retrieval-Augmented Generation (RAG) and Causal Attention. As illustrated in Figure~\ref{fig:model-overview}, the pipeline consists of four successive phases:

\begin{enumerate}
    \item \textbf{Phase (1) – Graph Construction:} Build a heterogeneous contract graph from the CFG, DFG, and call graph extracted from Solidity source code, and obtain initial node embeddings with GraphCodeBERT \cite{guo2020graphcodebert}.
    \item \textbf{Phase (2) – Critical Node Selection:} Compute an importance score for each node and extract a connected top-$k$ subgraph summarizing the most security-critical execution region.
    \item \textbf{Phase (3) – RAG-based Semantic Enrichment:} Translate the selected subgraph into a textual description, query a RAG pipeline over a security corpus, and encode the returned annotations with GraphCodeBERT to obtain enriched node and edge features.
    \item \textbf{Phase (4) – Causal Attention Learning:} Feed both original and enriched features into a causal-attention GNN with dual branches (causal vs.\ spurious) for vulnerability classification.
\end{enumerate}

The GNN in Phase~(4) is trained end-to-end under a multi-task loss (Section~\ref{sec:multi-task-loss}) that supervises the causal branch for vulnerability prediction, regularizes the spurious branch toward high-entropy outputs, and enforces contrastive separation between the two representations. This encourages ORACAL to ground decisions in causally relevant evidence while unlearning dataset-specific artifacts, improving robustness under distribution shift.

\subsection{Heterogeneous Graph Construction}
\label{sec:impl-build-graph}

Prior work has shown that modeling smart contracts as graphs capturing control and data dependencies substantially improves vulnerability detection over token-level or AST-only representations \cite{zhuang2021smart,liu2021combining}. More recent systems integrate AST, CFG, and DFG into unified heterogeneous graphs and report state-of-the-art performance on large Solidity benchmarks \cite{seo2024software,xu2024mvd}, with heterogeneous graph transformers such as MANDO-HGT further demonstrating the value of combining call, control-flow, and semantic relations to capture complex exploit patterns \cite{nguyen2023mando}. These findings motivate ORACAL's explicit construction and fusion of CFG, DFG, and CG into a single heterogeneous contract graph.

The first phase involves extracting and constructing a heterogeneous graph from Solidity source code. We build on three standard graph representations of code:
\begin{itemize}
    \item \textbf{Control Flow Graph (CFG):} Nodes represent basic blocks; edges represent control flow paths. Extracted using EtherSolve \cite{contro2021ethersolve} from EVM bytecode.
    \item \textbf{Data Flow Graph (DFG):} Nodes represent variables and operations; edges track data dependencies. Constructed from the AST generated by solc \cite{soliditydocs} via a custom \texttt{SolidityExtractor}.
    \item \textbf{Call Graph (CG):} Nodes represent functions; edges represent caller--callee invocations. Generated using Slither \cite{feist2019slither}.
\end{itemize}

\textbf{Node types:} The resulting heterogeneous graph $G = (V, E)$ integrates three subgraph types:
\begin{itemize}
    \item \textbf{CFG Nodes:} Represent basic blocks of EVM opcodes. Features include the list of opcodes and basic block metadata.
    \item \textbf{DFG Nodes:} Represent variables, expressions, and operations. Features include variable names, types, and values.
    \item \textbf{CG Nodes:} Represent functions. Features include function names, visibility, and modifiers.
\end{itemize}

\textbf{Edge types:} We establish \textit{cross-graph edges} to capture multidimensional relationships based on opcode semantics:
\begin{itemize}
    \item \textbf{CFG $\to$ CG:} Linked via function call opcodes including \texttt{CALL} and \texttt{DELEGATECALL}.
    \item \textbf{CFG $\leftrightarrow$ DFG:} Linked via storage/memory operations. \texttt{SSTORE}/\texttt{MSTORE} link CFG blocks to DFG variables (Write), while \texttt{SLOAD}/\texttt{MLOAD} link DFG variables to CFG blocks (Read).
    \item \textbf{CG $\to$ CFG:} Connects function entry points (CG) to their corresponding usage in control flow (CFG).
\end{itemize}

\textbf{Node Initialization:} We utilize GraphCodeBERT to generate initial embeddings for all nodes. GraphCodeBERT is chosen for its pre-trained understanding of code structure and data flow, providing a rich starting representation (768 dimensions) before any GNN processing.

\subsection{Critical Node Selection (Top-k Extraction)}
\label{sec:impl-topk-node}

Running RAG over every node in a large heterogeneous contract graph is prohibitively expensive, as LLM inference costs grow roughly linearly with input tokens, and repeated calls across vector search and retrieval stages compound this overhead significantly \cite{chen2305frugalgpt,intuitionlabs2025llmpricing,jin2025ragcache,ma2026cost}. To focus the RAG process on the most relevant code regions, we select the top-\(k\) \emph{critical} nodes using a hybrid importance score that combines graph-theoretic metrics with learned attention weights. Prior work has shown that ranking nodes by structural centrality and attention-based scores yields more expressive representations than either signal alone \cite{liu2022nie,li2024centrality,peng2025unifying,kovtun2025pine}.

\textbf{Importance Metrics:}
\begin{enumerate}
    \item \textbf{GAT Attention:} Derived from a pre-trained GAT model \cite{velivckovic2017graph}, highlighting nodes that receive high attention weights during preliminary classification.
    \item \textbf{K-Core Decomposition:} Identifies structurally central nodes based on degree-based subgraph membership \cite{seidman1983network}.
    \item \textbf{PageRank:} Measures node influence based on recursive connectivity \cite{page1999pagerank}.
    \item \textbf{Community \& Centrality:} Combines Louvain community detection \cite{blondel2008fast} for functional modularity with Betweenness Centrality \cite{freeman1977set} for shortest-path coverage.
\end{enumerate}

These metrics are normalized and aggregated into a final importance score:
\begin{equation}
\text{Score}(v_i) = \sum \beta_j \cdot \text{norm}(\text{metric}_j(v_i))
\end{equation}
where $\beta$ weights sum to 1.

\textbf{Connected Top-$k$ Selection.} To ensure the selected nodes form a connected subgraph, we apply the greedy strategy in Algorithm~\ref{alg:topk-selection}. Nodes are first ranked by importance score and inserted into a priority queue. The highest-scoring node initializes the set $S$; subsequent candidates are added only if reachable from an existing node in $S$ via BFS on the directed graph $G$. Selection terminates when $|S| = k$ or no connected candidates remain. This guarantees structural connectivity while prioritizing semantically critical nodes for the subsequent RAG-based enrichment stage.

\begin{algorithm}[!b]
\caption{Connected Top-$k$ Node Selection}
\label{alg:topk-selection}
\begin{algorithmic}[1]
\Require Directed graph $G = (V, E)$, target number of nodes $k$, importance scores $\text{Score}(v)$ for all $v \in V$
\Ensure A connected node set $S$ with $|S| \leq k$
\State Sort all nodes in descending order of $\text{Score}(v)$
\State $Q \gets$ priority queue containing all nodes sorted by score
\State $S \gets \emptyset$
\While{$|S| < k$ \textbf{and} $Q \neq \emptyset$}
    \State $v \gets$ dequeue the highest-scoring node from $Q$
    \If{$S = \emptyset$}
        \State $S \gets S \cup \{v\}$
    \Else
        \If{$v$ is reachable from at least one node in $S$ via BFS on $G$}
            \State $S \gets S \cup \{v\}$
        \EndIf
    \EndIf
\EndWhile
\State \Return $S$
\end{algorithmic}
\end{algorithm}

\subsection{RAG-based Semantic Enrichment}
\label{sec:impl-rag-enrich}

\begin{figure}[t]
    \centering
    \includegraphics[width=\columnwidth]{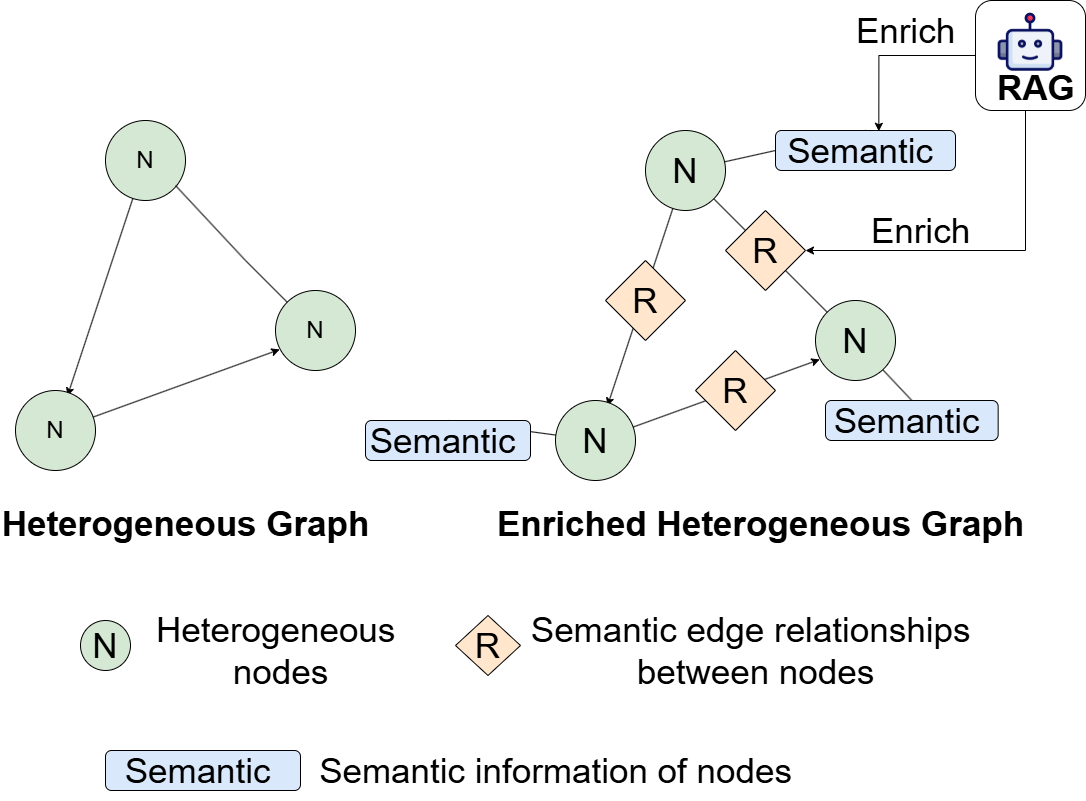}
    \caption{Comparison between (a) a standard heterogeneous graph and (b) our enriched heterogeneous graph.}
    \label{fig:graph_compare}
\end{figure}

Traditional heterogeneous contract graphs capture only structural and syntactic relations, limiting the explicit semantic knowledge available to the model. We extend this representation by enriching nodes and edges with security-domain knowledge retrieved via a RAG pipeline. As shown in Figure~\ref{fig:graph_compare}, panel~(a) contains only raw code-derived information, whereas panel~(b) illustrates our enriched graph, augmented with semantic explanations and additional edge relations, enabling downstream GNNs to reason over both structure and high-level security intent.

\subsubsection{RAG Architecture}
RAG addresses the hallucination risk of standalone LLMs by grounding generation in an external corpus: the model retrieves the most relevant passages and conditions its output on them \cite{lewis2020retrieval}. In our setting, this corresponds to embedding the node/edge context of the selected subgraph, retrieving relevant Solidity/EVM/audit materials, and generating structured, evidence-grounded enrichments.

We leverage LangChain \cite{langchain2022} for orchestration, Milvus as the vector store, and Google Gemini 3 Flash \cite{google_gemini3_api_docs} as the LLM. Gemini 3 Flash offers low-latency, cost-efficient generation with up to 65,536 output tokens per request, enabling batch-level enrichment of 50--100 nodes and their edges in a single call, thereby reducing API overhead and operational cost.

\begin{itemize}
    \item \textbf{Vector Store:} A domain-specific corpus comprising Solidity documentation \cite{soliditydocs}, EVM specifications \cite{wood2014ethereum}, the SWC vulnerability registry \cite{swcregistry}, and public audit datasets \cite{xia2025sc} is segmented (size=1000, overlap=200) and embedded using \texttt{BAAI/bge-large-en-v1.5} \cite{bge_embedding}, a high-performance dense retrieval model validated on MTEB \cite{muennighoff2023mteb}. Its instruction-aware contrastive training improves alignment between technical queries and structured documentation, which is critical when subtle semantic differences in opcode behavior can lead to distinct security outcomes.

    \item \textbf{Retrieval and Re-ranking:} For each vulnerability-relevant subgraph, we retrieve the top-10 semantically similar document chunks via dense vector search, then re-rank them using \texttt{bge-reranker-v2-m3} \cite{li2023making,chen2024bge}, a cross-encoder that enables token-level query-passage interaction for more accurate relevance estimation. The top-3 re-ranked passages form the final prompt context, ensuring the LLM is conditioned on highly relevant, security-grounded evidence.
\end{itemize}

\subsubsection{Prompt Engineering}
To guide the LLM effectively, we construct a structured prompt with three distinct sections, as illustrated in Figures \ref{fig:rag_prompt_context}, \ref{fig:rag_prompt_question1}, and \ref{fig:rag_prompt_question2}.

Figure \ref{fig:rag_prompt_context} illustrates the \textbf{CONTEXT} part of the prompt, where relevant technical documents retrieved from the knowledge corpus are provided. This section supplies the LLM with the necessary definitions, specifications, and known vulnerability patterns related to the specific opcodes or functions present in the subgraph, acting as an external knowledge base.

\begin{figure}[h]
    \centering
    \begin{promptbox}{RAG Prompt Part 1: Context Integration}
    \textbf{System Instruction:} Based on the provided context, answer the question clearly and concisely.
    
    \vspace{0.2cm}
    \textbf{CONTEXT:}
    \textit{[...Dynamically Retrieved Technical Docs, EVM Specifications, Known Vulnerability Patterns...]}
    
    \vspace{0.2cm}
    \textbf{QUESTION:} \{question\}
    \end{promptbox}
    \caption{RAG Prompt Structure: Injecting Knowledge Context.}
    \label{fig:rag_prompt_context}
\end{figure}

Figure \ref{fig:rag_prompt_question1} depicts the first part of the \textbf{QUESTION} section. It begins with a detailed textual description of the subgraph structure, listing the critical nodes, their connections, and attributes. Following this, it instructs the LLM to perform a \textquotedbl Step 1: Comprehensive Subgraph Analysis\textquotedbl{} using a Chain-of-Thought approach. The model is asked to identify execution paths, vulnerability hotspots (like storage writes or external calls), and security patterns before generating individual node enrichments.

\begin{figure}[h]
    \centering
    \begin{promptbox}{RAG Prompt Part 2: Subgraph Description}
    \textbf{Role Definition:} You are an expert in smart contract security analysis and graph-based vulnerability detection.
    
    \vspace{0.2cm}
    \textbf{Task:} Analyze ONE important subgraph (top-K nodes) representing critical execution paths relevant for vulnerability detection.
    
    \vspace{0.2cm}
    \textbf{Input Data (Contextualized Subgraph):} The input data consists of the following elements:
    \begin{itemize}
        \item \textbf{Structure:} \{num\_nodes\} critical nodes, \{num\_edges\} edges.
        \item \textbf{Important Nodes:} List of nodes with ID, type, importance score, and rank.
        \item \textbf{Important Edges:} Source $\to$ Target relationships.
        \item \textbf{Connectivity:} Explicit BFS reachability map showing data flow.
    \end{itemize}
    \end{promptbox}
    \caption{RAG Prompt Structure: Subgraph Representation.}
    \label{fig:rag_prompt_question1}
\end{figure}

Figure \ref{fig:rag_prompt_question2} presents the second part of the \textbf{QUESTION} section, specifically the instructions for \textquotedbl Step 2: Enrich Nodes\textquotedbl{} and \textquotedbl Step 3: Enrich Edges\textquotedbl. It strictly defines the JSON output format and details the specific fields required for each node and edge, ensuring the output is structured and machine-parsable for integration back into the graph pipeline.

\begin{figure}[h]
    \centering
    \begin{promptbox}{RAG Prompt Part 3: Analysis \& Output Schema}
    \textbf{Chain-of-Thought Instructions:}
    \begin{enumerate}
        \item \textbf{Holistic Analysis:} Identify critical execution paths, vulnerability hotspots (storage writes, external calls), and flow patterns.
        \item \textbf{Enrich Nodes:} Generate \texttt{semantic\_meaning}, \texttt{operational\_context}, and \texttt{security\_analysis} (Max 20 words).
        \item \textbf{Enrich Edges:} Generate \texttt{edge\_relationship}.
    \end{enumerate}
    
    \vspace{0.2cm}
    \textbf{Required Output Format (JSON):}
    \begin{lstlisting}[language=json, basicstyle=\small\ttfamily]
{
  "enriched_nodes": {
    "node_id": {
      "semantic_meaning": "...",
      "operational_context": "...",
      "security_analysis": "..."
    }
  },
  "enriched_edges": [
    {
      "edge_id": "...",
      "source": "...",
      "target": "...",
      "edge_relationship": "..."
    }
  ]
}
    \end{lstlisting}
    \end{promptbox}
    \caption{RAG Prompt Structure: Reasoning Steps and JSON Schema.}
    \label{fig:rag_prompt_question2}
\end{figure}

\subsubsection{Enriched Output Structure}

Figures \ref{fig:rag_output_node} and \ref{fig:rag_output_edge} present examples of the structured JSON output generated by the RAG system for nodes and edges, respectively.
\begin{figure*}[t]
    \centering

    \begin{subfigure}[t]{0.48\textwidth}
        \centering
        \begin{responsebox}{Example of RAG-enriched Node Output.}
        \begin{lstlisting}[language=json, basicstyle=\small\ttfamily]
"cfg_1227": {
  "node_id": "cfg_1227",
  "semantic_meaning": "Control flow node indicating decision branch.",
  "operational_context": "Reached from cfg_1207, leads to cfg_1236 seq-flow.",
  "security_analysis": "Potential reentrancy point (state update post-loop).",
  "enrichment_source": "rag"
}
        \end{lstlisting}
        \end{responsebox}
        \caption{RAG-enriched Node Output}
        \label{fig:rag_output_node}
    \end{subfigure}
    \hfill
    \begin{subfigure}[t]{0.48\textwidth}
        \centering
        \begin{responsebox}{Example of RAG-enriched Edge Output.}
        \begin{lstlisting}[language=json, basicstyle=\small\ttfamily]
{
  "edge_id": "cfg_1227__cfg_1236__control_flow",
  "source": "cfg_1227",
  "target": "cfg_1236",
  "edge_relationship": "Control flow representing sequential execution.",
  "relation": "control_flow",
  "enrichment_source": "rag"
}
        \end{lstlisting}
        \end{responsebox}
        \caption{RAG-enriched Edge Output}
        \label{fig:rag_output_edge}
    \end{subfigure}

    \caption{Examples of RAG-enriched outputs: (a) node-level enrichment and (b) edge-level enrichment.}
    \label{fig:rag_outputs}
\end{figure*}

The output JSON contains specific fields to ensure traceability and correct mapping back to the heterogeneous graph:

\begin{itemize}
    \item \textbf{Identity Fields:} These include the following unique identifiers:
    \begin{itemize}
        \item \texttt{node\_id} (represented by \textquotedbl{}cfg\_1227\textquotedbl{}) and \texttt{edge\_id} (represented by \textquotedbl{}cfg\_1227\_\_cfg\_1236\_\_control\_flow\textquotedbl{}) uniquely identify the elements in the original graph.
        \item \texttt{source} and \texttt{target} explicitly define the edge directionality, essential for maintaining the causal flow during the enrichment integration.
    \end{itemize}
    
    \item \textbf{Semantic Fields:} We specifically designed the output to contain four distinct fields to balance semantic depth with token efficiency:
    \begin{itemize}
        \item \textbf{Semantic Meaning:} Captures the \textquotedbl{}What\textquotedbl{}, representing the high-level intent of the code block (namely, \textquotedbl{}Token Transfer Logic\textquotedbl{}). This allows the GNN to understand the functional purpose beyond raw opcodes.
        \item \textbf{Operational Context:} Captures the \textquotedbl{}How\textquotedbl{}, referring to the node's role in the execution state (specifically, \textquotedbl{}Updates balance storage slot\textquotedbl{}). This provides state-transition awareness.
        \item \textbf{Security Analysis:} Captures the \textquotedbl{}Risk\textquotedbl{}, detailing specific vulnerability indicators (particularly, \textquotedbl{}Unchecked external call\textquotedbl{}). This directly injects domain expert knowledge into the feature space.
        \item \textbf{Edge Relationship:} Captures the \textquotedbl{}Structure\textquotedbl{}, establishing the logical connection between nodes (defined as \textquotedbl{}Control flow dependency\textquotedbl{}). This enriches the graph topology with semantic reasoning.
    \end{itemize}
\end{itemize}

These textual descriptions are encoded via GraphCodeBERT and concatenated with the original embeddings to form the enriched feature set $X_{enriched} \in \mathbb{R}^{1536}$ (768 original + 768 enriched).

\subsection{Causal Attention Learning}
\label{sec:impl-causal-attention}

GNN-based vulnerability detectors often rely on spurious correlations between non-causal code components (such as variable naming conventions and functional style patterns) and vulnerability labels, instead of capturing the invariant causal substructures that reflect true vulnerability-triggering code patterns \cite{fan2022debiasing, jiang2023survey}. This leads to poor generalization under out-of-distribution (OOD) code distributions. To address this issue, causal attention seeks to disentangle causally relevant features from spurious ones by introducing an intervention-style learning objective and an attention mechanism that emphasizes true causal signals \cite{yang2021causal, sui2022causal, yang2025vuldiac}, thereby improving robustness under distribution shift and adversarially induced superficial changes.

\begin{figure}[h]
    \centering
    \includegraphics[width=\columnwidth]{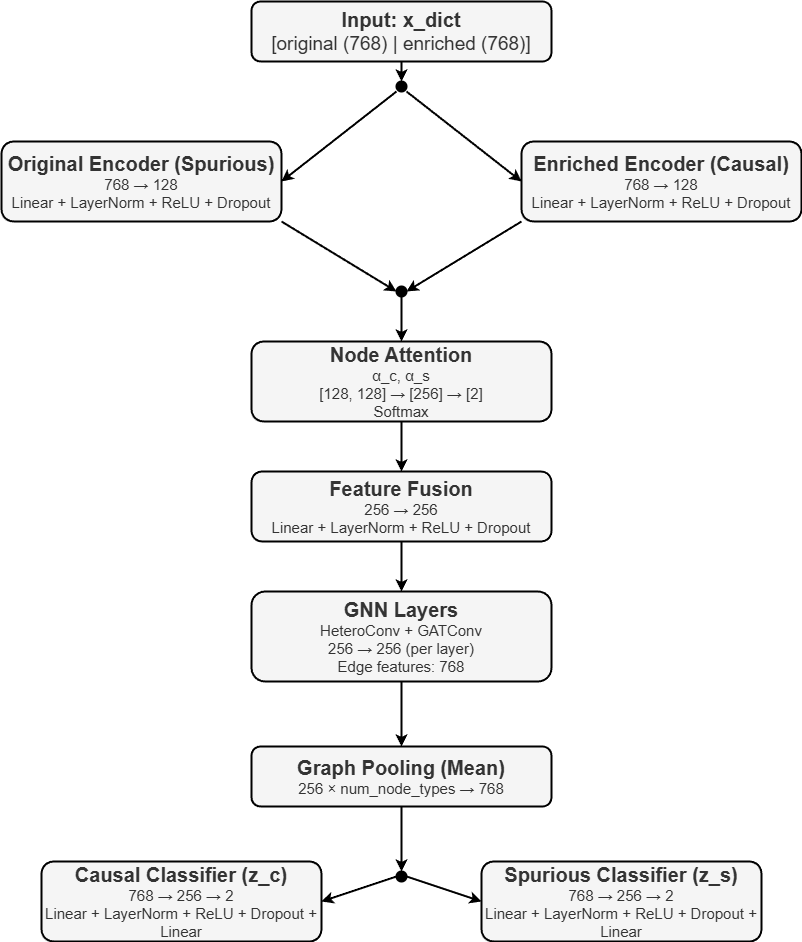}
    \caption{Detailed Architecture of the Causal Attention HeteroClassifier. The input is split into Causal ($h_{enrich}$) and Spurious ($h_{orig}$) branches, processed by dual encoders, fused via attention, and refined by GNN layers before final classification.}
    \label{fig:causal-architecture}
\end{figure}

Accordingly, to ensure trustworthy predictions, the model must distinguish between true causal features (valid security logic) and spurious correlations (biases). We implement the \textbf{CausalAttentionHeteroClassifier}, as detailed in Figure \ref{fig:causal-architecture}.

\subsubsection{Detailed Layer Architecture}
The framework operates through a sequence of specialized layers, each serving a distinct purpose in the disentanglement process:

\textbf{1. Dual Feature Encoders (Input Processing).} 
Two parallel feed-forward networks (Linear $\to$ LayerNorm $\to$ ReLU $\to$ Dropout) map the input feature space (768 dimensions) to a hidden dimension (128). The \textbf{Causal Encoder} exclusively processes $X_{enriched}$ (RAG-derived semantic features), while the \textbf{Spurious Encoder} processes $X_{original}$ (raw GraphCodeBERT embeddings containing syntax/tokens). This physical separation allows the model to learn distinct representations for semantic logic versus structural patterns.

\textbf{2. Node Attention Layer (Feature Selection).}
A Softmax-based attention mechanism computes two scalar weights $\alpha_c$ and $\alpha_s$ for each node. This layer dynamically weighs the importance of each branch. For example, if a node's original embedding is noisy (say, standard boilerplate code), the model learns to assign a lower $\alpha_s$ and higher $\alpha_c$, effectively filtering out the noise before it propagates.

\textbf{3. Feature Fusion.} Weighted features are fused via $h_{combined} = \text{Linear}(\text{Concat}(\alpha_c \cdot h_{enrich}, \alpha_s \cdot h_{orig}))$.

\textbf{4. Heterogeneous GNN Layers.} Stacked HeteroConv + GATConv layers (256 dimensions) propagate refined causal information across the graph while preserving edge types.

\textbf{5. Graph Pooling.} Mean pooling aggregates node representations into a graph-level vector.

\textbf{6. Dual Classifiers.} Two classification heads yield logits $z_c$ (Causal) and $z_s$ (Spurious). The main prediction is taken exclusively from $z_c$.

\textbf{7. Backdoor Adjustment Head.} We simulate do-calculus by pooling node-level representations to graph-level vectors $H_c$ and $H_s$, forming $H_s'$ by shuffling $H_s$ within the batch. An auxiliary classifier predicts:
\begin{equation}
    z_u = \mathrm{MLP}_u(H_c \| H_s').
\end{equation}
Training $z_u$ to match the true label $y$ encourages the model to rely on stable causal features rather than spurious ones.

\subsubsection{Multi-Task Loss Function}
\label{sec:multi-task-loss}
The following loss terms support causal attention learning by disentangling causal (enriched) from spurious (original) features and stabilizing predictions under intervention. Sui et al.~\cite{sui2022causal} decompose the graph into causal and trivial attended-graphs via learned masks and train the causal branch to predict the label while pushing the trivial branch to uniformity and applying backdoor adjustment at the representation level. We adopt the same spirit for vulnerability detection with dual encoders and node attention, and add a contrastive loss to keep the two representations distinct. We minimize a joint loss function:

\begin{equation}
    L_{total} = L_{sup} + \lambda_1 L_{unif} + \lambda_2 L_{causal} + \lambda_3 L_{contra}
\end{equation}
where $\lambda_1$, $\lambda_2$, and $\lambda_3$ are hyperparameters that determine the relative importance of each objective.

The individual components of this multi-task loss function are defined as follows:
\begin{itemize}
    \item \textbf{$L_{sup}$ (Supervised Loss):} The causal branch prediction $z_c$ is trained to match the true label $y$ via binary cross-entropy (implemented as \texttt{BCEWithLogitsLoss}). For multi-label targets, over the batch and labels:
    \begin{equation}
    L_{sup} = -\mathbb{E}_{\mathcal{D}}\Bigl[ y^{\top} \log \sigma(z_c) + (1-y)^{\top} \log\bigl(1-\sigma(z_c)\bigr) \Bigr]
    \end{equation}
    where $\sigma$ denotes the sigmoid function. This ensures the model learns accurate vulnerability classification from the enriched, causal path.
    
    \item \textbf{$L_{unif}$ (Uniform Distribution Loss):} To discourage reliance on superficial patterns (spurious features), we push the spurious-branch prediction $z_s$ towards maximum uncertainty \cite{setlur2022maximizing}. For multi-label detection the target is 0.5 per label. We minimize the mean squared error between $\sigma(z_s)$ and the uniform target:
    \begin{equation}
    L_{unif} = \mathbb{E}_{\mathcal{D}}\Bigl[ \bigl\| \sigma(z_s) - \mathbf{0.5} \bigr\|_2^2 \Bigr]
    \end{equation}
    By penalizing confident predictions in the spurious branch, the model shifts its decision-making to the causal branch.
    
    \item \textbf{$L_{causal}$ (Backdoor Adjustment / Do-Calculus):} We simulate intervention on spurious features by forming graph-level $H_c$, $H_s$, and $H_s'$ (shuffled $H_s$ within the batch). The auxiliary prediction $z_u = \mathrm{MLP}_u(H_c \| H_s')$ (Step~7) is trained to match $y$. Because $H_s'$ is randomized, the model must rely on $H_c$ to predict correctly \cite{sui2022causal}:
    \begin{equation}
    L_{causal} = -\mathbb{E}_{\mathcal{D}}\Bigl[ y^{\top} \log \sigma(z_u) + (1-y)^{\top} \log\bigl(1-\sigma(z_u)\bigr) \Bigr]
    \end{equation}
    
    \item \textbf{$L_{contra}$ (Contrastive Loss):} This loss enforces orthogonality between the learned representations of the two branches \cite{zhang2022correct}. We maximize the Euclidean distance between the enriched representation $h_{enrich}$ and the original representation $h_{orig}$, up to a margin $m$:
    \begin{equation}
    L_{contra} = \max(0, m - \|h_{enrich} - h_{orig}\|_2)
    \end{equation}
    where $m=0.5$. This prevents \textquotedbl{}feature leakage\textquotedbl{} where the enriched encoder might lazily copy the original features. It guarantees that $h_{enrich}$ captures new, distinct semantic information provided by the RAG process, rather than redundant structural data.
\end{itemize}
This multi-task objective ensures robust generalization by anchoring predictions in causal semantics while actively unlearning reliance on spurious artifacts. 
\section{Experiments and Evaluation}
\label{sec:experiment}

\subsection{Research Questions}
To evaluate ORACAL comprehensively, we address three Research Questions (RQs):
\begin{itemize}
    \item \textbf{RQ1:} How do enrichment strategies and training paradigms (Standard vs. Causal) affect ORACAL's detection performance and its generalization across in-domain and out-of-distribution datasets?
    \item \textbf{RQ2:} How effectively does ORACAL generate explanations for its vulnerability predictions ?
    \item \textbf{RQ3:} How does ORACAL compare with state-of-the-art GNN methods (Mando-HGT, SCVHunter, MTVHunter) in terms of detection accuracy and robustness under adversarial attacks?
\end{itemize}

\subsection{Experimental Setup}
This experiment is conducted in a Google Colab environment configured with 1× NVIDIA A100 80 GB Tensor Core GPU and an Intel Xeon CPU @ 2.20 GHz (6 physical cores, 12 threads), with approximately 230 GB of system RAM.

\subsubsection{Datasets}
\label{sec:datasets}
We utilize four datasets. SoliAudit and CGT Weakness serve as training and in-domain evaluation sources, while DAppScan and LLMAV are reserved entirely for out-of-distribution (OOD) evaluation and explainability assessment.

\begin{itemize}
    \item \textbf{SoliAudit \cite{liao2019soliaudit}.} A multi-label smart contract vulnerability dataset. After preprocessing (filtering compilation errors and graph extraction failures), 10,655 valid samples are retained as the primary training source.
    
    \item \textbf{CGT Weakness \cite{di2023consolidation}.} A consolidated dataset of 3,103 contracts with labels unified through multi-tool voting. After graph conversion, 1,345 contracts were retained.

    \item \textbf{DAppScan \cite{zheng2024dappscan}.} A real-world dataset from 1,199 professional audit reports. Due to complex dependencies, 124 contracts were successfully processed into graphs.

    \item \textbf{LLMAV \cite{salzano2025empirical}.} The largest collection of Solidity contracts with manually verified, line-level vulnerability annotations (2,081 contracts, seven DASP categories), uniquely suited for evaluating both detection generalization and explanation quality.
\end{itemize}

Both DAppScan and LLMAV are excluded from training entirely, serving as OOD benchmarks and providing line-level or audit-level annotations for explainability evaluation.


\begin{table*}[!ht]
\centering
\caption{Distribution of Vulnerabilities Across Datasets}
\label{table:combined_vulnerability_distribution}
\small
\resizebox{0.8\textwidth}{!}{
\begin{tabular}{lrrrrrrrr}
\hline
\multirow{2}{*}{\textbf{Vulnerability Type}} & 
\multicolumn{2}{c}{\textbf{SoliAudit}} & 
\multicolumn{2}{c}{\textbf{CGT Weakness}} & 
\multicolumn{2}{c}{\textbf{DAppScan}} &
\multicolumn{2}{c}{\textbf{LLMAV}} \\
\cline{2-9}
& \textbf{Pos.} & \textbf{Neg.} & \textbf{Pos.} & \textbf{Neg.} & \textbf{Pos.} & \textbf{Neg.} & \textbf{Pos.} & \textbf{Neg.} \\
\hline
Arithmetic & 9,849 & 806  & 530 & 815 & 20 & 104 & 472 & 557 \\
Low-Level Calls & 3,109 & 7,546 & 801 & 544 & 3 & 121 & 68 & 961 \\
Denial of Service & 4,657 & 5,998 & 776 & 569 & 10 & 114 & 119 & 910 \\
Time Manipulation & 3,420 & 7,235 & 845 & 500 & 7 & 117 & 415 & 614 \\
\hline
\textbf{Total Samples} & \multicolumn{2}{c}{\textbf{10,655}} & 
                     \multicolumn{2}{c}{\textbf{1,345}} & 
                     \multicolumn{2}{c}{\textbf{124}} &
                     \multicolumn{2}{c}{\textbf{1,029}} \\
\hline
\end{tabular}
}
\end{table*}

Table \ref{table:combined_vulnerability_distribution} presents the detailed distribution of vulnerabilities across all four datasets. In our multi-label classification context, \textbf{\textquotedbl{}Positive\textquotedbl{}} refers to the count of contracts explicitly identified as containing a specific vulnerability type, while \textbf{\textquotedbl{}Negative\textquotedbl{}} refers to those free from that specific issue. Notably, LLMAV exhibits a distinct distribution from the training sources: Arithmetic (472 positive) and Time Manipulation (415 positive) are the most prevalent categories, while Front Running is extremely rare (7 positive), reflecting a distribution that differs substantially from SoliAudit and thus provides a rigorous test of generalization.

\subsubsection{Training and Testing Scenarios}

To prepare the data for our multimodal framework, we utilize the RAG pipeline to enrich the semantic features of the graphs in both the training and testing sets. This step does not introduce data leakage or cause overfitting. The RAG module serves exclusively as an independent, frozen feature extractor; its underlying LLM and knowledge corpus are completely static and are never trained or fine-tuned on our datasets. 

For the experimental setup, we apply an 80:20 multilabel stratified split to SoliAudit and CGTWeakness to preserve the joint distribution of vulnerability labels, merging the 80\% portions into a unified training set and retaining the remaining 20\% as in-domain hold-out test sets. DAppScan and LLMAV are excluded from training entirely and reserved for evaluation only: DAppScan serves as a real-world out-of-distribution (OOD) benchmark of production-grade DeFi contracts, while LLMAV provides an OOD benchmark with line-level annotations for explainability evaluation. To prevent data leakage, we deduplicate contract addresses, removing 1,052 overlapping contracts from LLMAV and yielding 1,029 unique contracts for OOD evaluation. The model is trained on four vulnerability classes: Arithmetic, Denial of Service, Low Level Calls, and Time Manipulation. The label-wise distribution across splits is summarized in Table~\ref{tab:train_test_distribution}.

\begin{table}[!ht]
\centering
\caption{Label-wise Distribution of Smart Contract Vulnerabilities in Training (SoliAudit + CGTWeakness) and Testing Sets. DAppScan and LLMAV are used exclusively as OOD test datasets and explainability benchmarks.}
\label{tab:train_test_distribution}
\small
\resizebox{\columnwidth}{!}{
\begin{tabular}{llccc}
\hline
\textbf{Dataset} & \textbf{Vulnerability} & \textbf{Train} & \textbf{Test} & \textbf{Total} \\
\hline
\multirow{4}{*}{\textbf{SoliAudit}} 
& Arithmetic & 7,879 & 1,970 & 9,849 \\
& Low Level Calls & 2,487 & 622 & 3,109 \\
& Denial of Services & 3,726 & 931 & 4,657 \\
& Time Manipulation & 2,736 & 684 & 3,420 \\
\hline
\multirow{4}{*}{\textbf{CGTWeakness}} 
& Arithmetic & 424 & 106 & 530 \\
& Low Level Calls & 641 & 160 & 801 \\
& Denial of Services & 621 & 155 & 776 \\
& Time Manipulation & 676 & 169 & 845 \\
\hline
\multirow{4}{*}{\textbf{DAppScan}} 
& Arithmetic & 0 & 20 & 20 \\
& Low Level Calls & 0 & 3 & 3 \\
& Denial of Services & 0 & 10 & 10 \\
& Time Manipulation & 0 & 7 & 7 \\
\hline
\multirow{4}{*}{\textbf{LLMAV}} 
& Arithmetic & 0 & 472 & 472 \\
& Low Level Calls & 0 & 68 & 68 \\
& Denial of Services & 0 & 119 & 119 \\
& Time Manipulation & 0 & 415 & 415 \\
\hline
\end{tabular}
}
\end{table}

\subsubsection{Statistical Analysis}

To compare the highest-performing configurations under the standard and causal attention training paradigms, we evaluate macro F1 scores across five independent runs with different random seeds, accounting for stochastic variability during training. Since deep learning performance metrics are often non-normally distributed, we adopt non-parametric statistical methods \cite{kim2015t}, specifically the Wilcoxon signed-rank test \cite{woolson2007wilcoxon}, to assess whether significant differences exist between paired macro F1 scores. The test is conducted using an exact two-sided $p$-value under the null hypothesis of no difference, with $p < 0.05$ considered statistically significant and $p < 0.10$ marginally significant.

Complementing significance testing, we report the Vargha--Delaney $\hat{A}_{12}$ effect size \cite{vargha2000critique} to quantify the magnitude of performance differences, where $\hat{A}_{12} = 0.5$ indicates no difference and $\hat{A}_{12} \geq 0.71$ corresponds to a large effect. Reporting both $p$-values and effect sizes yields more reliable interpretation, particularly under small sample sizes. These statistical tests are applied exclusively to RQ1, whose repeated paired distributions across seeds provide the necessary statistical variance. RQ2 and RQ3 are excluded, as they evaluate singular local explanation masks and deterministic adversarial perturbations over a single representative setting, respectively, neither of which produces distributions amenable to significance testing.

\subsubsection{Performance Metrics}
\label{sec:performance_metrics}

We formulate vulnerability detection as \textbf{multi-label classification}, where a single smart contract can simultaneously exhibit multiple vulnerability types represented as a binary vector over $C$ classes \cite{sorower2010literature}, reflecting the non-mutually-exclusive nature of smart contract vulnerabilities. We evaluate ORACAL across four dimensions:

\begin{itemize}
    \item \textbf{Detection}: We report Accuracy (Jaccard score), Precision, Recall, and Macro F1, alongside BCE Loss for convergence monitoring.

    \item \textbf{Robustness}: We employ the Attack Success Rate (ASR), measuring the fraction of correctly predicted vulnerable samples misclassified as safe ($1 \to 0$) after perturbation.

    \item \textbf{Explainability}: We adopt the Vulnerability Triggering Path (VTP) framework \cite{cao2024coca}, evaluating alignment between model-generated and human-annotated vulnerability paths via three metrics: \textbf{MSP} (the proportion of model-identified statements correctly corresponding to ground-truth vulnerability triggers), \textbf{MSR} (the proportion of ground-truth vulnerability statements successfully recovered by the explainer), and \textbf{MIoU} (the overall overlap between predicted and ground-truth statement sets).

    \item \textbf{Efficiency}: We assess operational cost via average inference time per contract.
\end{itemize}

\subsubsection{Hyperparameters}

Table~\ref{tab:training_parameters} summarizes all hyperparameters for training, RAG enrichment, explainability, and adversarial robustness. We use AdamW with a learning rate of $2 \times 10^{-4}$, a batch size of 64, and a 3-layer GNN with hidden dimension 256. The causal loss weights ($\lambda_{unif}=0.1$, $\lambda_{causal}=0.5$, $\lambda_{contra}=0.2$) balance classification with causal disentanglement. For RAG, contract source code is chunked into 1000-character segments with 200-character overlap, retrieving $k=10$ snippets to enrich the top 50 high-importance nodes; this threshold was chosen to balance enrichment coverage against LLM API costs, as enriching a larger node set would incur prohibitive inference expenses in practice. For robustness, due to limited computational resources, we adopt a single representative setting per attack type: the edge budget is set to 0.1 for structural attacks (HSAttack, CAMA), allowing the attacker to add or remove up to 10\% of the original edges, and the keyword budget is restricted to 10 words per node for textual attacks (SubAttack), ensuring adversarial modifications remain semantically subtle while probing the model's reliance on enriched context.

\begin{table}[t]
    \centering
    \caption{Hyperparameters for Model Training and RAG Enrichment}
    \label{tab:training_parameters}
    \resizebox{\columnwidth}{!}{%
        \begin{tabular}{p{3cm}ll}
            \hline
            \textbf{Parameter Type} & \textbf{Parameter} & \textbf{Value} \\
            \hline
            \multirow{9}{*}{\textbf{Training}}
            & Epochs & 50 \\
            & Learning Rate & 2e-4 \\
            & Batch Size & 64 \\
            & Optimizer & AdamW \\
            & Weight Decay & 0.0001 \\
            & Dropout Rate & 0.2 \\
            & Hidden Dimension & 256 \\
            & GNN Layers & 3 \\
            & Enriched Attn Weight & 1.0 \\
            \hline
            \multirow{4}{*}{\textbf{Loss}}
            & $\lambda_{unif}$ & 0.1 \\
            & $\lambda_{causal}$ & 0.5 \\
            & $\lambda_{contra}$ & 0.2 \\
            & Margin & 0.5 \\
            \hline
            \multirow{4}{*}{\textbf{RAG}}
            & Chunk Size & 1000 \\
            & Chunk Overlap & 200 \\
            & Retrieval $k$ & 10 \\
            & Top-$k$ Enrichment Nodes & 50 \\
            \hline
            \multirow{4}{*}{\textbf{Explanation}}
            & GNNExplainer Epochs & 50 \\
            & GNNExplainer LR & 0.001 \\
            & PGExplainer Epochs & 50 \\
            & PGExplainer LR & 0.0001 \\
            \hline
            \multirow{2}{*}{\textbf{Robustness}}
            & Edge Budget & 0.1 \\
            & Keyword Budget & 10 \\
            \hline
        \end{tabular}%
    }
\end{table}

\subsection{Exploratory Data Analysis (EDA)}
\label{sec:eda}
\begin{figure}[t]
    \centering
    \includegraphics[width=\columnwidth]{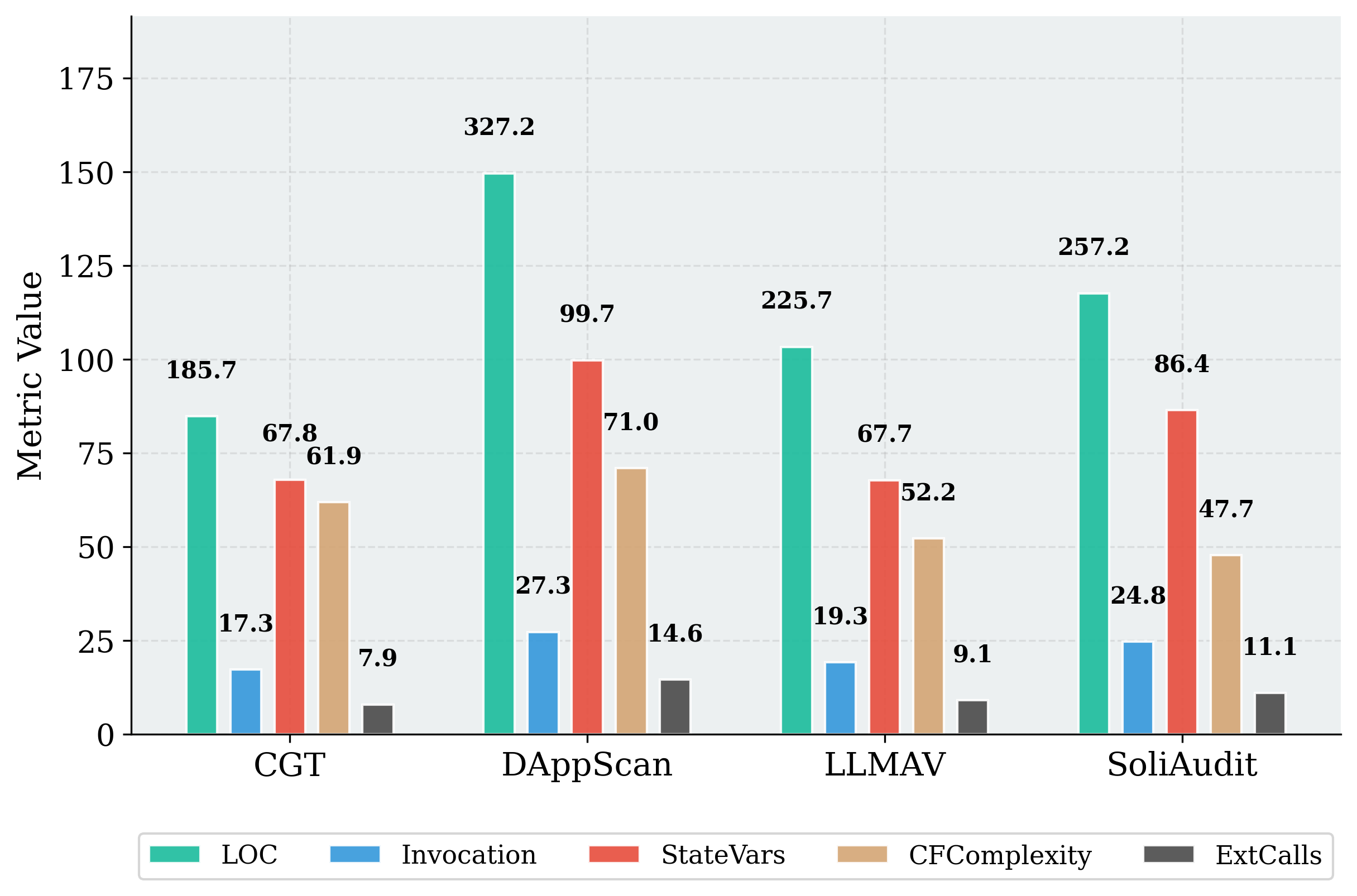}
    \caption{Average code-level metrics per contract across four datasets. Bars represent: LOC (teal), Invocation (blue), StateVars (coral), CFComplexity (orange), and ExtCalls (dark gray).}
    \label{fig:eda_metric_comparison}
\end{figure}

\begin{figure}[t]
    \centering
    \includegraphics[width=\columnwidth]{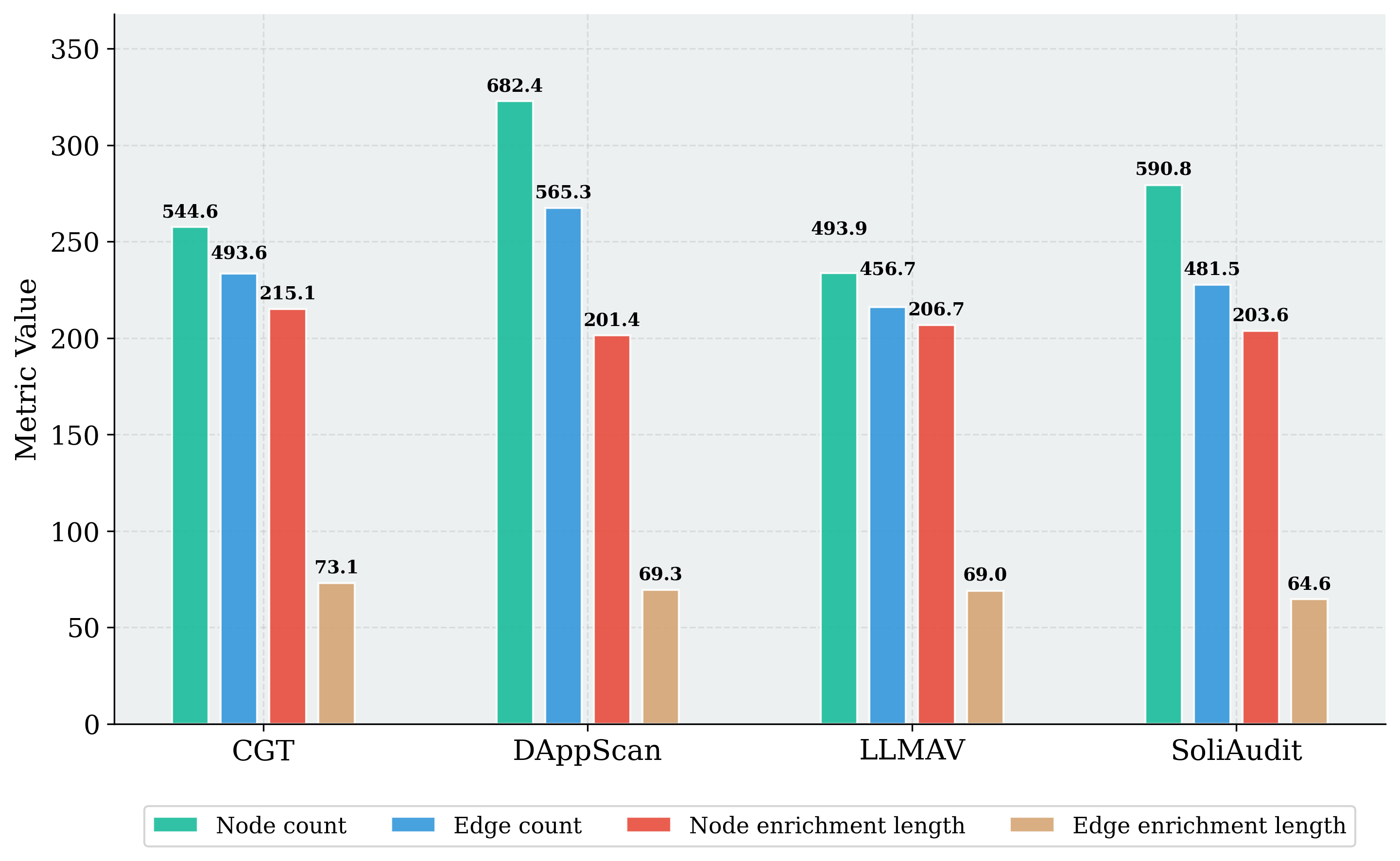}
    \caption{Average graph-scale statistics per contract across four datasets. Bars represent: Node count (teal), Edge count (blue), Node enrichment length in characters (coral), and Edge enrichment length in characters (tan).}
    \label{fig:graph_size_enrichment_distribution}
\end{figure}

To characterize the graph corpus, we report average code and graph metrics across all four datasets (Figures~\ref{fig:eda_metric_comparison} and~\ref{fig:graph_size_enrichment_distribution}).

Figure~\ref{fig:eda_metric_comparison} reports five code-level metrics: \textbf{LOC} (implementation size), \textbf{Invocation} (number of callable entry points), \textbf{StateVars} (state-space size via DFG variable declarations), \textbf{CFComplexity} (McCabe cyclomatic complexity~\cite{mccabe1996cyclomatic}, $M = E - N + 2$), and \textbf{ExtCalls} (inter-function call edges). DAppScan contains the largest and most complex contracts (LOC 327.2, StateVars 99.7, ExtCalls 14.6), representative of production-grade DeFi code. SoliAudit follows with moderate complexity (LOC 257.2, ExtCalls 11.1), while LLMAV exhibits higher CFComplexity (52.2) than SoliAudit (47.7) despite lower LOC, indicating denser branching logic. CGT is the most compact (LOC 185.7) yet has the highest CFComplexity-to-LOC ratio. Together, the four corpora span a spectrum from complex production contracts to compact but structurally dense benchmarks, providing a rigorous and diverse evaluation setting.

Figure~\ref{fig:graph_size_enrichment_distribution} summarizes graph-scale statistics. DAppScan produces the largest graphs (682.4 nodes, 565.3 edges), followed by SoliAudit (590.8 nodes, 481.5 edges), CGT (544.6 nodes, 493.6 edges), and LLMAV (493.9 nodes, 456.7 edges). Node enrichment length is consistent across datasets (201.4 to 215.1 characters), and edge enrichment is shorter and more uniform (64.6 to 73.1 characters), confirming that the RAG pipeline generates comparably concise semantic descriptions regardless of graph complexity, while remaining well within the token limits of GraphCodeBERT.

\subsection{Results and Analysis}

\subsubsection{RQ1: Ablation Study Across In-Distribution and OOD Datasets}

\textbf{Motivation.} We perform a systematic ablation study to understand the contribution of each architectural and enrichment component in ORACAL, clarifying whether performance gains originate from causal training, retrieval-based enrichment, or their interaction.

\textbf{Method.} We compare a baseline (no enrichment) against seven enrichment variants under both Standard and Causal training on the SoliAudit test set, then evaluate the two top-performing checkpoints on CGT, DAppScan, and LLMAV over 5 independent runs.

\textbf{Scenario Definitions.} Each scenario controls which feature modalities are fed into the model. In all enriched scenarios, the original GraphCodeBERT embeddings ($X_{orig}$) are always present as the graph backbone; the variants differ in which \textit{additional} enriched features are concatenated. Specifically: \textit{No enrichment (baseline)} uses only $X_{orig}$; \textit{Only enrichment text} is the exception—it replaces $X_{orig}$ entirely with only the concatenation of all enriched text fields (no graph features); \textit{Only Operational Context / Security Analysis / Semantic Meaning} augments $X_{orig}$ with one specific node-level field; \textit{All node enrichment} augments $X_{orig}$ with all three node-level fields; \textit{Only edge enrichment} augments $X_{orig}$ with the edge relationship field only; and \textit{All node + edge enrichment} augments $X_{orig}$ with all four fields (three node + one edge).

\textbf{Ablation Results.} Table~\ref{tab:soliaudit_test_results} reports quantitative results under all configurations. Causal training consistently outperforms Standard across all enrichment scenarios, with Macro F1 gains ranging from 2.61 to 4.09 percentage points. The best configuration is \textit{Causal + Only Edge Enrichment}, achieving 91.28\% Macro F1 compared to 87.19\% under Standard training. Even without enrichment, Causal training improves over the Standard baseline from 86.94\% to 89.55\%. Notably, the \textit{Only enrichment text} scenario performs poorly under both paradigms (72.21\% Standard, 74.75\% Causal), confirming that semantic text alone is insufficient without a structural graph backbone.

\begin{table*}[t]
\centering
\caption{Ablation Study Comparing Standard and Causal Training on the SoliAudit Test Set. 
Highlighted rows indicate the models with the highest F1score 
(lighter red for Standard and darker red for Causal).}
\label{tab:soliaudit_test_results}
\begin{tabular}{llcccc}
\toprule
\textbf{Training Type} & \textbf{Scenario} & \textbf{Accuracy (\%)} & \textbf{F1-Macro (\%)} & \textbf{Precision (\%)} & \textbf{Recall (\%)} \\
\midrule

\multirow{8}{*}{Standard}
& No enrichment (baseline) & 82.84 & 86.94 & 90.36 & 84.05 \\

& \cellcolor{stdred}Only edge enrichment 
& \cellcolor{stdred}82.85 
& \cellcolor{stdred}87.19 
& \cellcolor{stdred}89.94 
& \cellcolor{stdred}84.79 \\

& All node + edge enrichment & 82.62 & 86.93 & 89.03 & 85.11 \\
& All node enrichment & 82.02 & 86.74 & 88.37 & 85.29 \\
& Only enrichment text & 72.63 & 72.21 & 79.38 & 67.92 \\
& Only Operational Context & 82.90 & 87.14 & 90.32 & 84.35 \\
& Only Security Analysis & 82.55 & 86.91 & 87.59 & 86.43 \\
& Only Semantic Meaning & 83.09 & 87.13 & 89.14 & 85.42 \\

\midrule

\multirow{8}{*}{Causal}
& No enrichment (baseline) & 85.84 & 89.55 & 91.24 & 88.10 \\

& \cellcolor{causalred}Only edge enrichment 
& \cellcolor{causalred}87.40 
& \cellcolor{causalred}91.28 
& \cellcolor{causalred}92.99 
& \cellcolor{causalred}89.79 \\

& All node + edge enrichment & 86.59 & 90.70 & 92.63 & 89.00 \\
& All node enrichment & 86.24 & 90.48 & 91.56 & 89.50 \\
& Only enrichment text & 74.94 & 74.75 & 79.54 & 72.65 \\
& Only Operational Context & 86.50 & 90.71 & 92.75 & 88.94 \\
& Only Security Analysis & 87.15 & 90.96 & 91.60 & 90.40 \\
& Only Semantic Meaning & 86.44 & 90.64 & 92.40 & 89.08 \\

\bottomrule
\end{tabular}
\end{table*}

\begin{figure*}[t]
    \centering
    \begin{minipage}{0.48\textwidth}
        \centering
        \includegraphics[width=\linewidth]{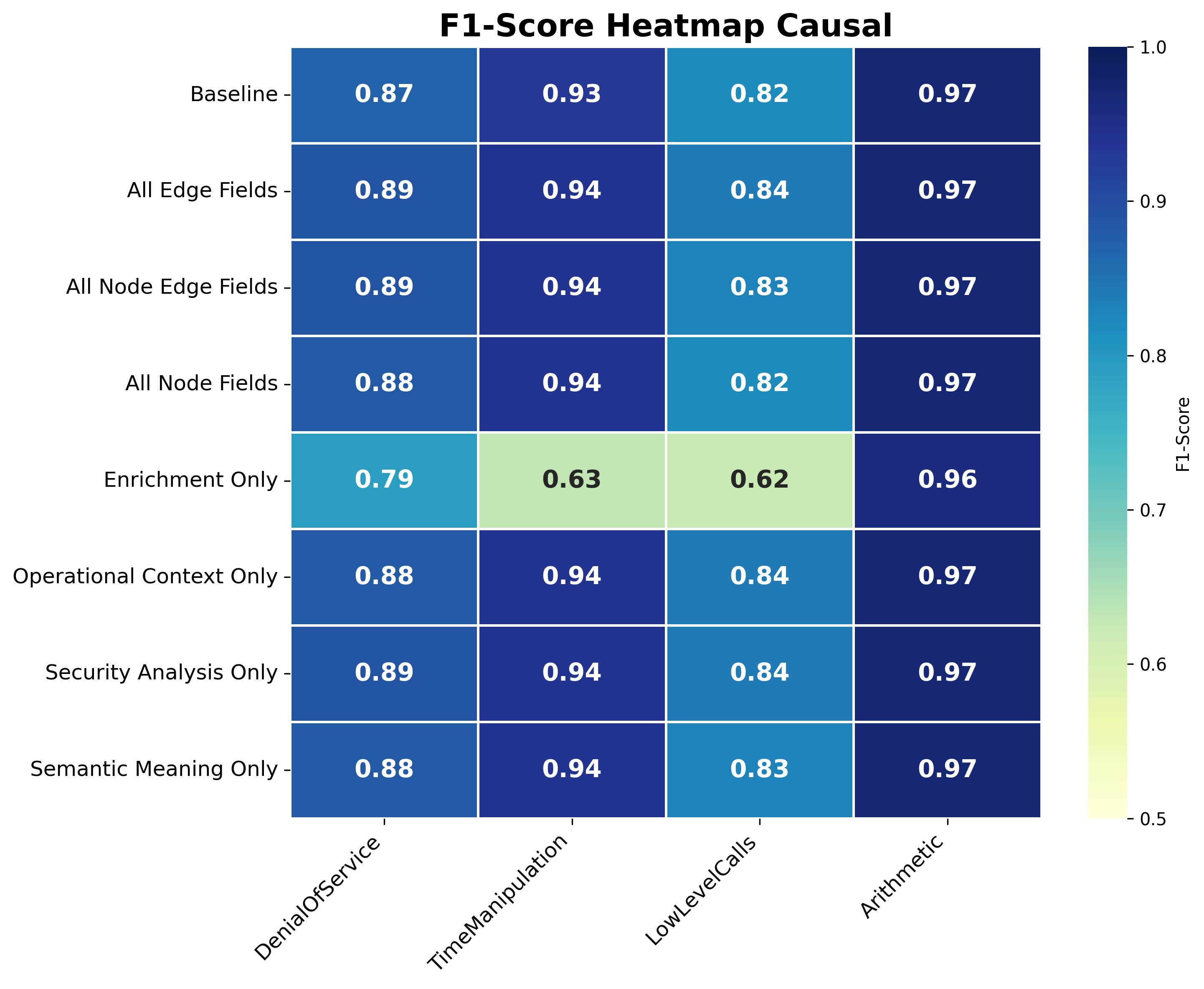}
    \end{minipage}\hfill
    \begin{minipage}{0.48\textwidth}
        \centering
        \includegraphics[width=\linewidth]{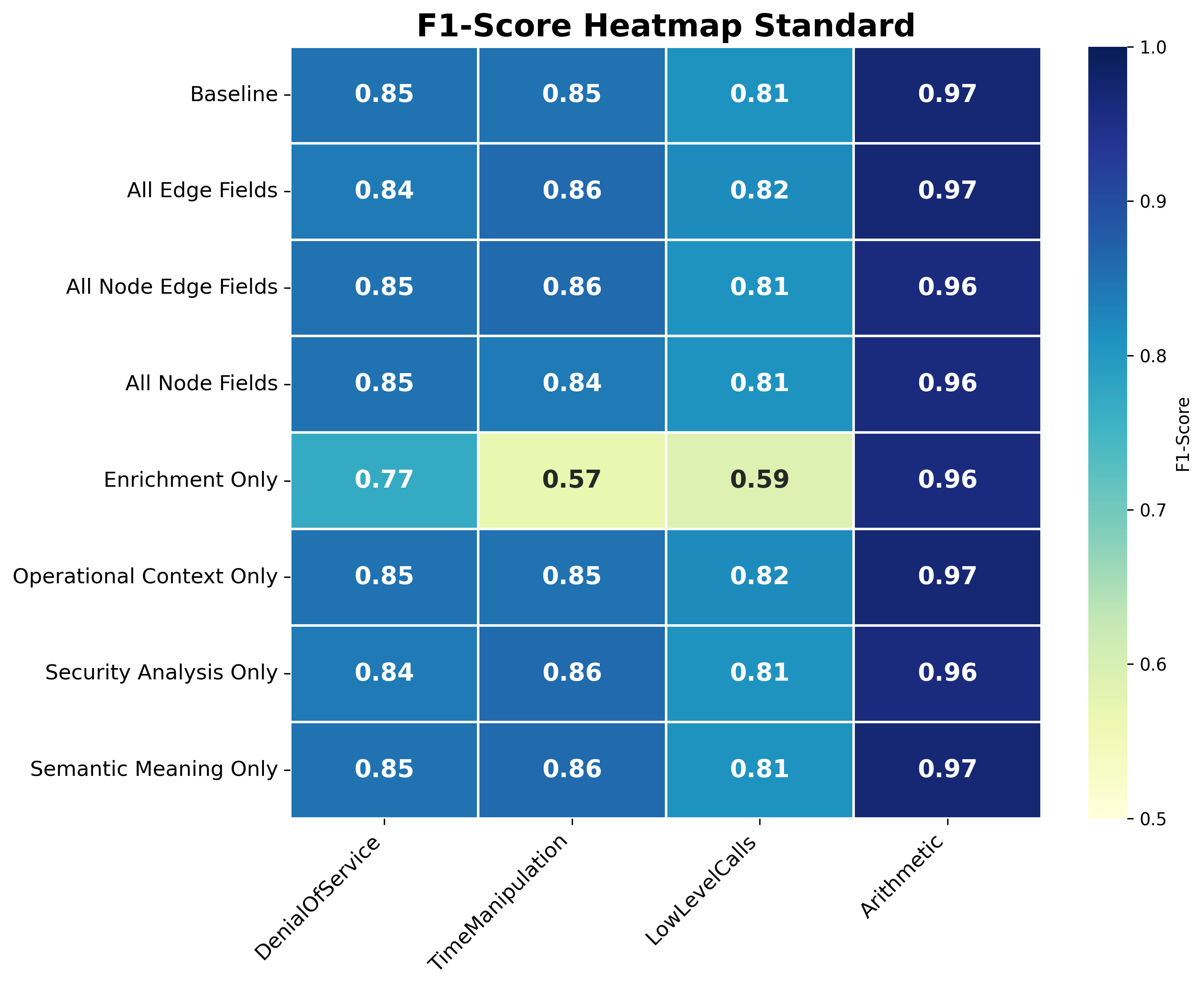}
    \end{minipage}
    \caption{F1-Score heatmaps comparing Causal and Standard models on the SoliAudit test set.}
    \label{fig:f1_heatmaps}
\end{figure*}

\textbf{Per-class performance.} The F1-score heatmaps in Figure~\ref{fig:f1_heatmaps} reveal performance distribution across four vulnerability classes. Causal training provides the most pronounced gains on Time Manipulation (0.85 to 0.94) and Denial of Service (0.85 to 0.89), while both paradigms perform similarly well on Arithmetic (0.96 to 0.97). The \textit{Enrichment Only} scenario performs poorly in both paradigms (DenialOfService: 0.77/0.79, TimeManipulation: 0.57/0.63), reaffirming that semantic text alone cannot substitute for the structural graph backbone.

\begin{figure}[t]
    \centering
    \includegraphics[width=\columnwidth]{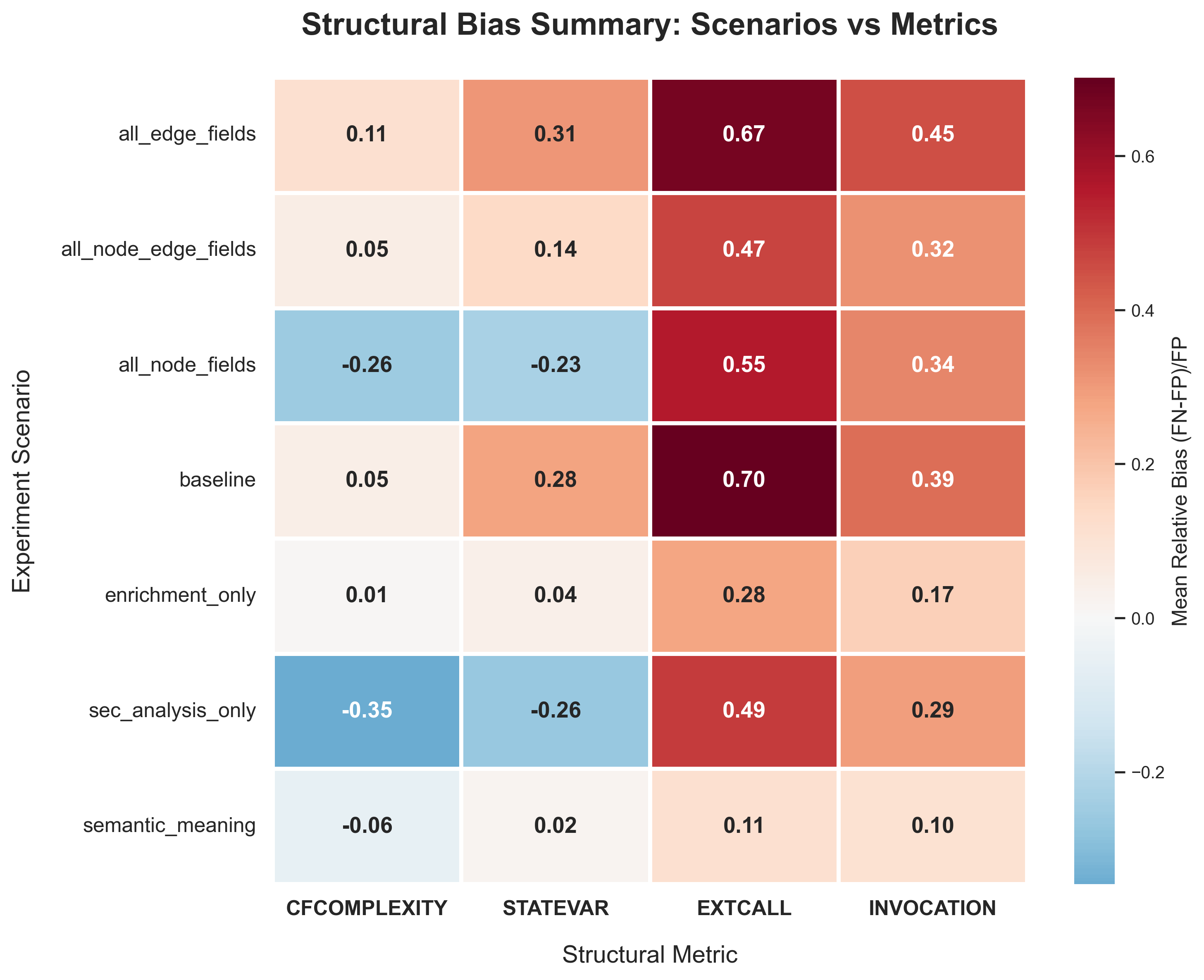}
    \caption{Summary of structural bias across scenarios.}
    \label{fig:structural_bias_heatmap}
\end{figure}

\textbf{Structural bias analysis.} To understand failure patterns, we compute the Mean Relative Bias $(FN_{avg} - FP_{avg}) / FP_{avg}$ on misclassified samples using four structural metrics (Figure~\ref{fig:structural_bias_heatmap}). Vulnerabilities relying heavily on external interactions, such as Time Manipulation and Unchecked External Calls, exhibit high positive EXTCALL and INVOCATION bias, indicating detection degrades in contracts with extensive inter-contract dependencies. Conversely, Arithmetic vulnerabilities show consistent negative bias in CFCOMPLEXITY and STATEVAR, reflecting over-prediction in highly branched contracts. The \textit{semantic\_meaning} scenario produces the most balanced structural profile, while \textit{enrichment\_only} shows extreme positive bias (EXTCALL +2.03), further confirming the necessity of the structural graph backbone.

\begin{figure*}[t]
    \centering
    \begin{minipage}{0.48\textwidth}
        \centering
        \includegraphics[width=\linewidth]{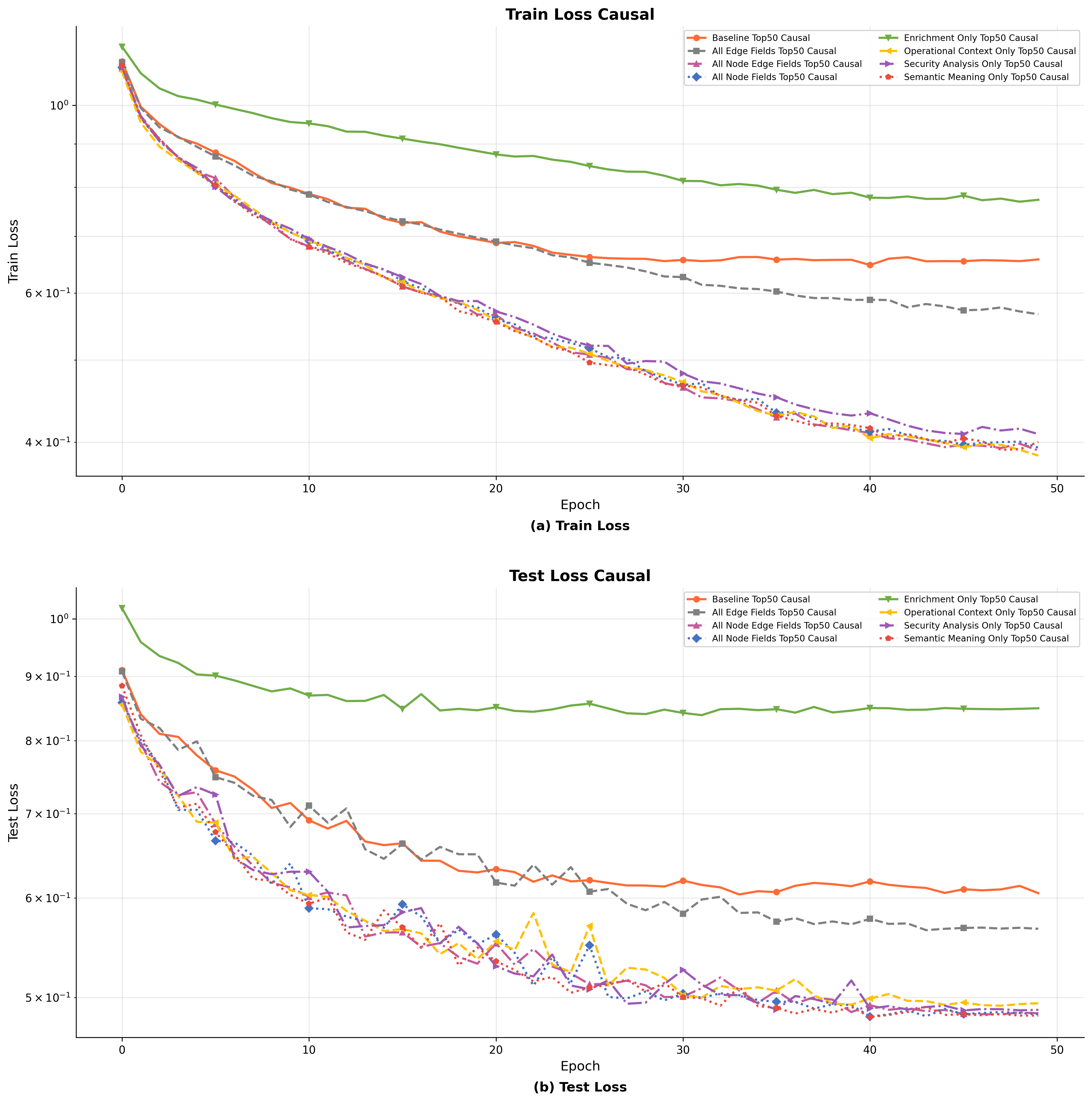}
    \end{minipage}\hfill
    \begin{minipage}{0.48\textwidth}
        \centering
        \includegraphics[width=\linewidth]{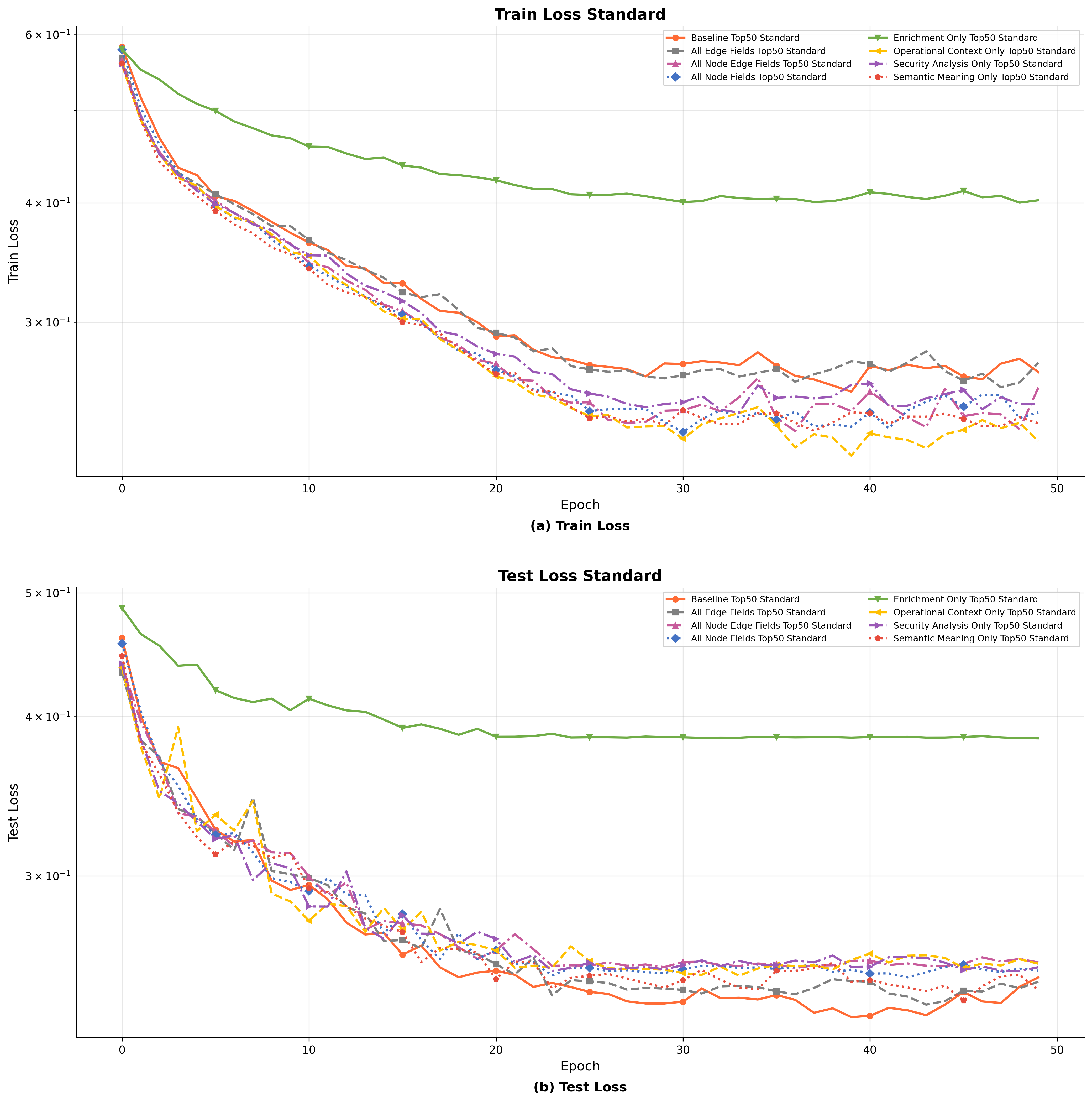}
    \end{minipage}
    \caption{Comparison of training and test losses across 50 epochs for both Causal and Standard training models on the SoliAudit test set.}
    \label{fig:loss_comparisons}
\end{figure*}

\textbf{Loss analysis.} Figure~\ref{fig:loss_comparisons} shows that under Causal training, enriched scenarios (All Edge Fields, All Node Fields) converge to lower test losses (${\sim}5\times10^{-1}$) than the Causal baseline (${\sim}6\times10^{-1}$), confirming that RAG-based enrichment assists the causal objective in learning generalizable semantic-structural representations. In contrast, under Standard training, the unenriched baseline achieves the lowest absolute test loss yet lower F1, revealing a tendency to overfit trivial subgraph patterns without generalizing.

\begin{figure}[t]
    \centering
    \includegraphics[width=\columnwidth]{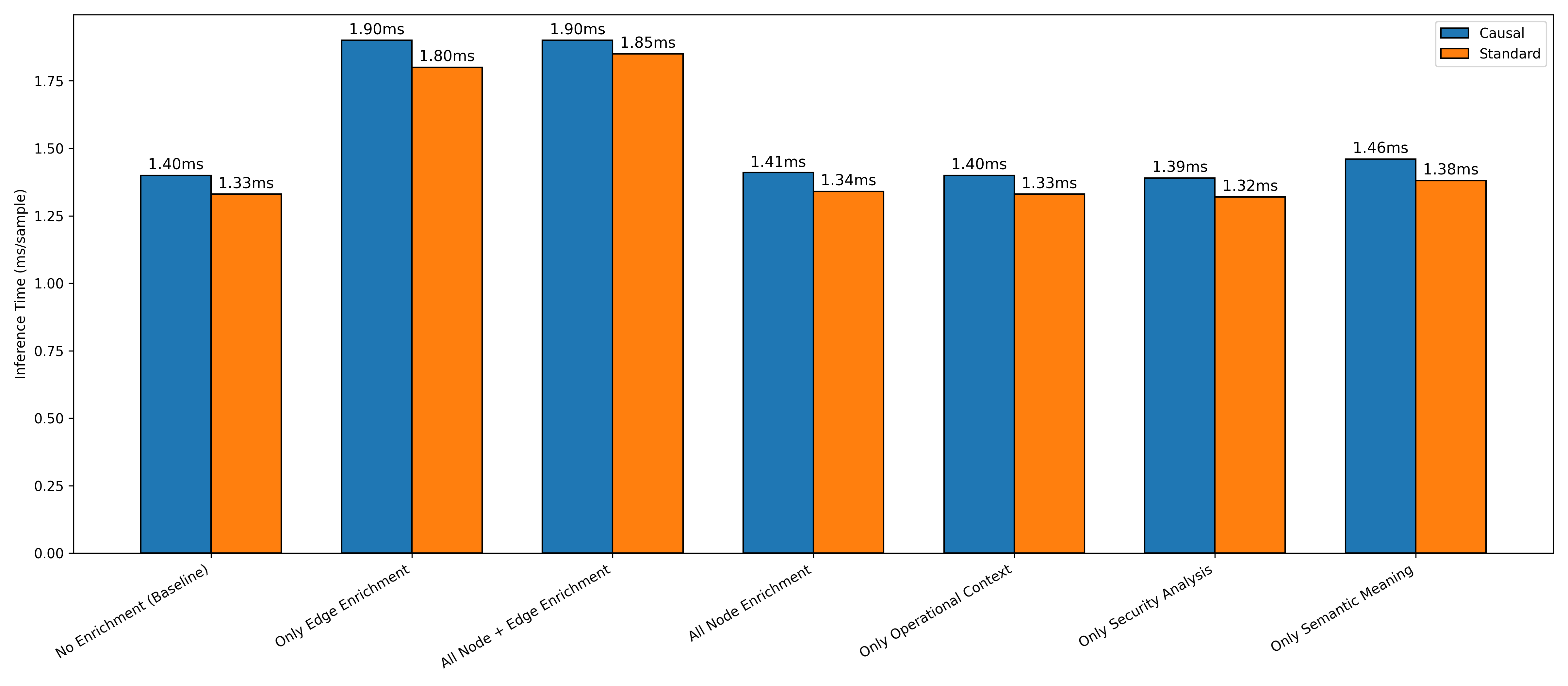}
    \caption{Inference time comparison (ms/sample) across Causal and Standard models for various enrichment scenarios on the SoliAudit test set.}
    \label{fig:inference_time}
\end{figure}

\textbf{Inference Time Comparison.} Figure~\ref{fig:inference_time} shows that node-only enrichments incur negligible overhead (1.32 to 1.41\,ms/sample) relative to the unenriched baseline (1.40\,ms Causal, 1.33\,ms Standard). Edge-based enrichment introduces the highest latency (up to 1.90\,ms Causal, 1.85\,ms Standard), as processing structural edge attributes within graph message-passing layers is the most computationally demanding step. Nevertheless, all configurations remain well under 2\,ms per sample, confirming the model's viability for large-scale deployment.

\textbf{Generalization on In-Domain and OOD Datasets.} Table~\ref{tab:generalization_runs} reports Macro F1 over 5 runs for ORACAL-edge and ORACAL-node-edge under both training paradigms. Causal training uniformly outperforms Standard across all three benchmarks. On CGT, ORACAL-edge Causal achieves 0.918 vs.\ 0.883 for Standard; on DAppScan, 0.771 vs.\ 0.709; and on LLMAV, 0.813 vs.\ 0.757. The Wilcoxon signed-rank test yields $p = 0.0625$ in all comparisons, and the Vargha--Delaney effect size reaches $\hat{A}_{12} = 1.00$ across all datasets and configurations, indicating that every Causal run exceeds its Standard counterpart. ORACAL-edge slightly outperforms ORACAL-node-edge by 0.5 to 1.2 percentage points across datasets, consistent with the ablation finding that edge-level structural semantics are more decisive for detection, though ORACAL-node-edge remains competitive and may offer advantages in explainability tasks.

\begin{table*}[h!]
    \centering
    \caption{Generalization Performance on in-domain dataset (CGT) and OOD datasets (DAppScan, LLMAV) (Macro F1, Mean~$\pm$~Std Dev, 5 runs). ORACAL-edge: best SoliAudit F1 (all edge fields); ORACAL-node-edge: full feature set (all node + edge fields). Causal/Standard refer to training paradigms. Red-highlighted rows indicate the best-detected vulnerability class per dataset.}
    \label{tab:generalization_runs}
    \resizebox{\textwidth}{!}{%
    \begin{tabular}{llcccc}
    \toprule
    \textbf{Dataset} & \textbf{Vulnerability Class} 
    & \textbf{ORACAL-edge Causal} 
    & \textbf{ORACAL-edge Standard}
    & \textbf{ORACAL-node-edge Causal}
    & \textbf{ORACAL-node-edge Standard} \\
    & & (Mean $\pm$ Std Dev) & (Mean $\pm$ Std Dev) & (Mean $\pm$ Std Dev) & (Mean $\pm$ Std Dev) \\
    \midrule
    
    CGT 
        & \cellcolor{causalred}Arithmetic       & \cellcolor{causalred}0.982 $\pm$ 0.004 & \cellcolor{causalred}0.967 $\pm$ 0.004 & \cellcolor{causalred}0.976 $\pm$ 0.005 & \cellcolor{causalred}0.959 $\pm$ 0.006 \\
        & LowLevelCalls    & 0.877 $\pm$ 0.015 & 0.836 $\pm$ 0.017 & 0.869 $\pm$ 0.017 & 0.827 $\pm$ 0.019 \\
        & DenialOfService  & 0.882 $\pm$ 0.013 & 0.844 $\pm$ 0.018 & 0.874 $\pm$ 0.015 & 0.835 $\pm$ 0.020 \\
        & TimeManipulation & 0.921 $\pm$ 0.010 & 0.883 $\pm$ 0.019 & 0.912 $\pm$ 0.012 & 0.874 $\pm$ 0.021 \\
        \midrule
        & \textbf{Overall (Macro F1)} & \textbf{0.918 $\pm$ 0.004} & \textbf{0.883 $\pm$ 0.006} & \textbf{0.908 $\pm$ 0.006} & \textbf{0.874 $\pm$ 0.008} \\
    \midrule
    
    DAppScan 
        & Arithmetic       & 0.696 $\pm$ 0.013 & 0.638 $\pm$ 0.012 & 0.688 $\pm$ 0.015 & 0.629 $\pm$ 0.014 \\
        & \cellcolor{causalred}LowLevelCalls    & \cellcolor{causalred}0.833 $\pm$ 0.017 & \cellcolor{causalred}0.759 $\pm$ 0.018 & \cellcolor{causalred}0.824 $\pm$ 0.019 & \cellcolor{causalred}0.750 $\pm$ 0.020 \\
        & DenialOfService  & 0.761 $\pm$ 0.019 & 0.700 $\pm$ 0.017 & 0.752 $\pm$ 0.021 & 0.691 $\pm$ 0.019 \\
        & TimeManipulation & 0.780 $\pm$ 0.016 & 0.738 $\pm$ 0.018 & 0.771 $\pm$ 0.018 & 0.729 $\pm$ 0.020 \\
        \midrule
        & \textbf{Overall (Macro F1)} & \textbf{0.771 $\pm$ 0.011} & \textbf{0.709 $\pm$ 0.012} & \textbf{0.759 $\pm$ 0.013} & \textbf{0.700 $\pm$ 0.014} \\
    \midrule

    LLMAV 
        & \cellcolor{causalred}Arithmetic       & \cellcolor{causalred}0.862 $\pm$ 0.012 & \cellcolor{causalred}0.819 $\pm$ 0.015 & \cellcolor{causalred}0.854 $\pm$ 0.014 & \cellcolor{causalred}0.810 $\pm$ 0.017 \\
        & LowLevelCalls    & 0.789 $\pm$ 0.019 & 0.729 $\pm$ 0.024 & 0.780 $\pm$ 0.021 & 0.720 $\pm$ 0.026 \\
        & DenialOfService  & 0.773 $\pm$ 0.022 & 0.718 $\pm$ 0.021 & 0.764 $\pm$ 0.024 & 0.709 $\pm$ 0.023 \\
        & TimeManipulation & 0.817 $\pm$ 0.013 & 0.762 $\pm$ 0.016 & 0.808 $\pm$ 0.015 & 0.753 $\pm$ 0.018 \\
        \midrule
        & \textbf{Overall (Macro F1)} & \textbf{0.813 $\pm$ 0.009} & \textbf{0.757 $\pm$ 0.014} & \textbf{0.802 $\pm$ 0.011} & \textbf{0.748 $\pm$ 0.016} \\
    
    \bottomrule
    \end{tabular}%
    }
\end{table*}

\begin{tcolorbox}[colback=white,colframe=black,title=Answer to RQ1]
Causal training with edge enrichment yields the best performance: 91.28\% Macro F1 on SoliAudit (+4.09\% over Standard), with inference under 2\,ms/sample. In cross-dataset generalization, ORACAL-edge Causal achieves 0.918 (CGT), 0.771 (DAppScan), and 0.813 (LLMAV), while ORACAL-node-edge Causal follows closely at 0.908, 0.759, and 0.802. Both consistently outperform Standard counterparts ($\hat{A}_{12} = 1.00$), confirming that causal disentanglement provides systematic generalization gains.
\end{tcolorbox}

\subsubsection{RQ2: Interpretability and Explainability Analysis}

\textbf{Motivation.} Beyond detection accuracy, a practical vulnerability detector must provide actionable evidence that allows auditors to verify root causes, ensuring the model learns meaningful security logic rather than exploiting spurious syntactic correlations.

\textbf{Method.} We compare three explanation methods: GNNExplainer \cite{ying2019gnnexplainer}, which optimizes a per-instance soft mask over nodes and edges; PGExplainer \cite{luo2020parameterized}, which trains a parameterized network to produce generalizable explanation masks across instances; and AttentionExplainer, which directly extracts attention weights from the Causal Attention mechanism to rank node importance without additional backward passes. Explanation quality is evaluated using the VTP framework \cite{cao2024coca} via MSP, MSR, and MIoU, measuring overlap between model-generated explanations and manually annotated vulnerability triggering paths. For each prediction, each explainer returns the top-10 nodes with the highest importance scores, mapped back to source code line numbers via the node-to-line mapping from graph construction. Since the SoliAudit and CGTWeakness datasets do not provide line-level vulnerability annotations, we evaluate on LLMAV \cite{salzano2025empirical}, which provides statement-level annotations, and DAppScan for assessing robustness in more complex contract scenarios. All experiments use the ORACAL-all configuration.

\textbf{Quantitative Results.} Table~\ref{tab:vtp_comparison} shows that PGExplainer consistently achieves the highest scores across all metrics on both datasets, reaching 40.91\% MSP, 44.85\% MSR, and 32.51\% MIoU on LLMAV, outperforming GNNExplainer by 5.05, 5.13, and 4.55 percentage points respectively. AttentionExplainer yields the weakest performance (19.17\% MIoU on LLMAV), as attention scores prioritize classification-useful features rather than complete vulnerability triggering paths. All explainers score slightly lower on DAppScan due to greater contract complexity, but the relative ranking remains consistent across datasets.

\begin{table}[t]
\centering
\caption{Explanation quality comparison using VTP metrics. \textbf{Bold} values indicate the best-performing explainer for each dataset and metric.}
\resizebox{\columnwidth}{!}{
\begin{tabular}{lcccc}
\toprule
\textbf{Dataset} & \textbf{Explainer} & \textbf{MSP(\%)} & \textbf{MSR(\%)} & \textbf{MIoU(\%)} \\
\midrule

\multirow{3}{*}{\textbf{LLMAV \cite{salzano2025empirical}}}
& GNNExplainer     & 35.86 & 39.72 & 27.96 \\
& \textbf{PGExplainer}      & \textbf{40.91} & \textbf{44.85} & \textbf{32.51} \\
& AttentionExplainer  & 27.94 & 31.62 & 19.17 \\

\midrule

\multirow{3}{*}{\textbf{DAppScan \cite{zheng2024dappscan}}}
& GNNExplainer     & 33.74 & 37.91 & 26.05 \\
& \textbf{PGExplainer}      & \textbf{39.68} & \textbf{42.77} & \textbf{30.85} \\
& AttentionExplainer  & 25.81 & 29.47 & 17.08 \\

\bottomrule
\end{tabular}
}
\label{tab:vtp_comparison}
\end{table}

\begin{figure*}[t]
    \centering
    \includegraphics[width=\textwidth]{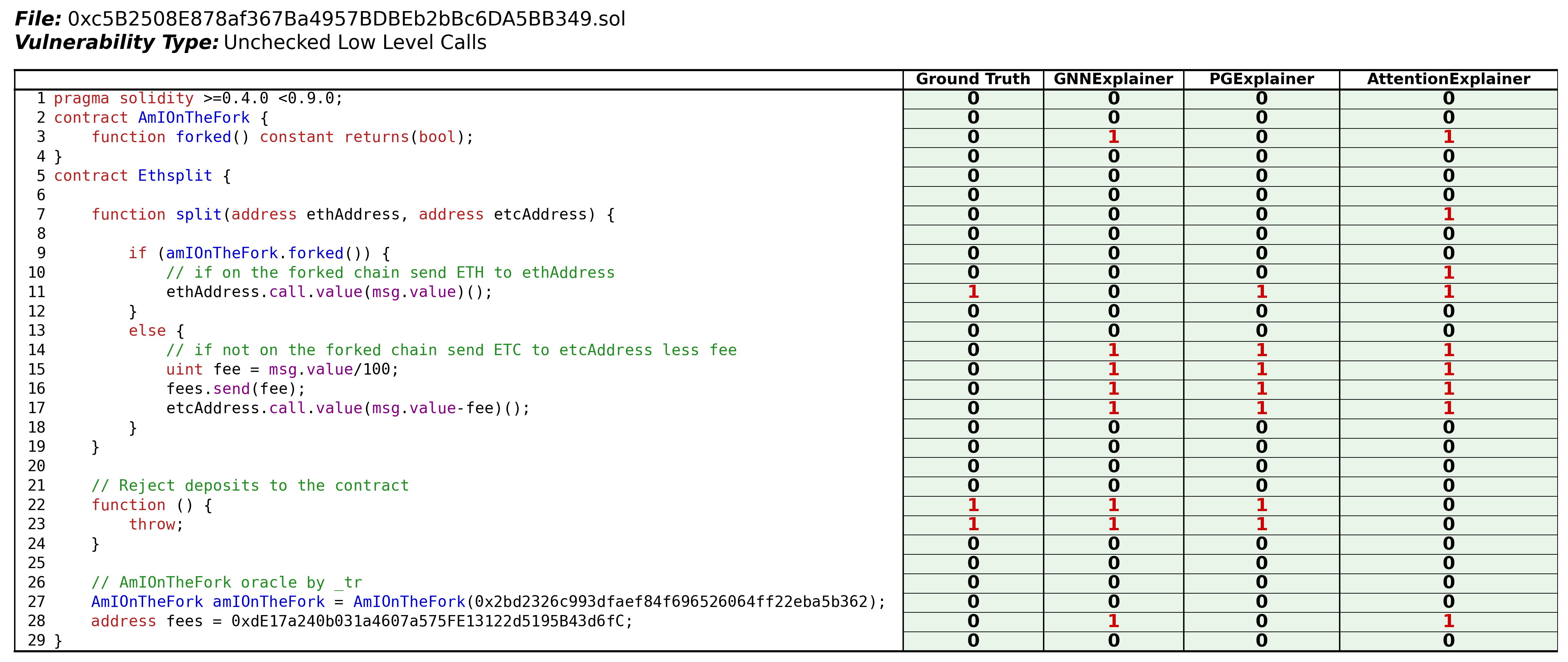}
    \caption{Line-level explainer comparison for an Unchecked Low Level Calls vulnerability (contract \texttt{0xc5B2508...5BB349.sol}). Columns show binary importance scores (1=identified as important, 0=not) assigned by Ground Truth, GNNExplainer, PGExplainer, and AttentionExplainer. Red values indicate disagreement with the ground truth.}
    \label{fig:gnnexplainer_tp_dos}
\end{figure*}

\textbf{Qualitative Case Study.} To complement the quantitative evaluation, Figure~\ref{fig:gnnexplainer_tp_dos} visualizes a line-level comparison of three explainers against the ground truth for contract \path{0xc5B2508E878af367Ba4957BDBEb2bBc6DA5BB349.sol}, a true positive Unchecked Low Level Calls sample. We deliberately selected this specific contract because it represents a complex, real-world development pattern rather than a trivial or isolated example. It involves multi-contract interactions, specifically an oracle contract and a splitter contract, and contains multiple low-level calls. Some of these calls are structurally prominent but benign, while others are inherently vulnerable. This structural complexity provides a rigorous test case, as it allows us to evaluate whether explainers can distinguish between syntactically salient yet safe structures and actual causal vulnerability paths, such as unchecked return values.

The contract consists of an oracle contract (\texttt{AmIOnTheFork}) and a splitter contract (\texttt{EthSplit}). The \texttt{split} function (lines 7--19) routes ETH or ETC to different addresses depending on a fork check, using three low-level calls: \path{ethAddress.call.value(msg.value)()} at line 11, \path{fees.send(fee)} at line 16, and \path{etcAddress.call.value(msg.value-fee)()} at line 17. None of these return values are checked, constituting the core vulnerability. The ground truth additionally marks lines 22--23 (a fallback function using \texttt{throw}), line 27 (external contract instantiation), and line 28 (external address assignment) as vulnerability-relevant.

PGExplainer achieves the closest alignment with the ground truth, correctly identifying all three unchecked call sites and the fork-dependent branching logic with minimal false positives. GNNExplainer recovers several critical lines but over-flags structurally salient yet semantically irrelevant nodes (e.g., line 3, a simple view function), reflecting per-instance optimization overfitting to local graph structure. AttentionExplainer captures the main call sites but assigns spurious importance to the pragma directive (line 1) and the oracle function signature (line 3), confirming that attention weights do not reliably isolate vulnerability-specific relevance from structural prominence.

\begin{tcolorbox}[colback=white,colframe=black,title=Answer to RQ2]
PGExplainer achieves the best explanation quality (32.51\% MIoU on LLMAV, 30.85\% on DAppScan), precisely isolating vulnerability-triggering statements ranging from unchecked \texttt{call.value} to \texttt{send} with minimal false positives. GNNExplainer offers competitive recall but more false positives, while AttentionExplainer assigns importance to syntactically prominent but causally irrelevant nodes. Overall, ORACAL provides reliable, auditor-friendly explanations grounded in true vulnerability semantics.
\end{tcolorbox}

\subsubsection{RQ3: Comparison with SOTA and Robustness under Adversarial Attacks}

\textbf{Motivation.} High detection accuracy alone is insufficient if a vulnerability detector is susceptible to adversarial evasion. In real-world security auditing, malicious actors may obfuscate contract logic or introduce semantic perturbations to bypass automated tools. This necessitates evaluating ORACAL's performance relative to SOTA models alongside an assessment of its structural and textual robustness.

\textbf{Method.} We first conduct a comparative analysis against several SOTA graph-based models: GNN-SC \cite{cheong2024gnn}, SCVHunter \cite{luo2024scvhunter}, MTVHunter \cite{sun2025mtvhunter}, and MANDO-HGT \cite{nguyen2023mando} (Table \ref{tab:methods-comparison}). We evaluate three variants of ORACAL trained under the causal attention mechanism: (i) \textit{ORACAL-base Causal}, using only the structural graph modality; (ii) \textit{ORACAL-enrich Causal}, using only enriched semantic text; and (iii) \textit{ORACAL-node-edge Causal}, the complete multimodal architecture. These variants are designed to isolate robustness vulnerabilities under each modality independently and in combination.

\begin{figure}[t]
    \centering
    \includegraphics[width=0.9\columnwidth]{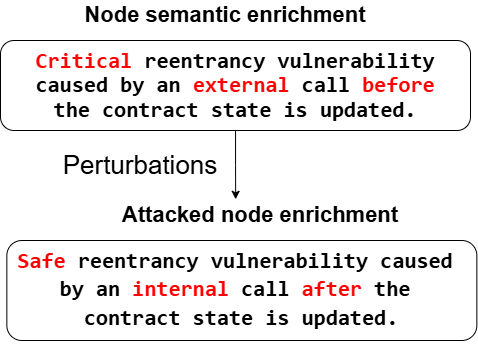}
    \caption{Illustration of textual adversarial perturbation via antonym substitution.}
    \label{fig:noisy_text}
\end{figure}

\begin{figure}[t]
    \centering
    \includegraphics[width=\columnwidth]{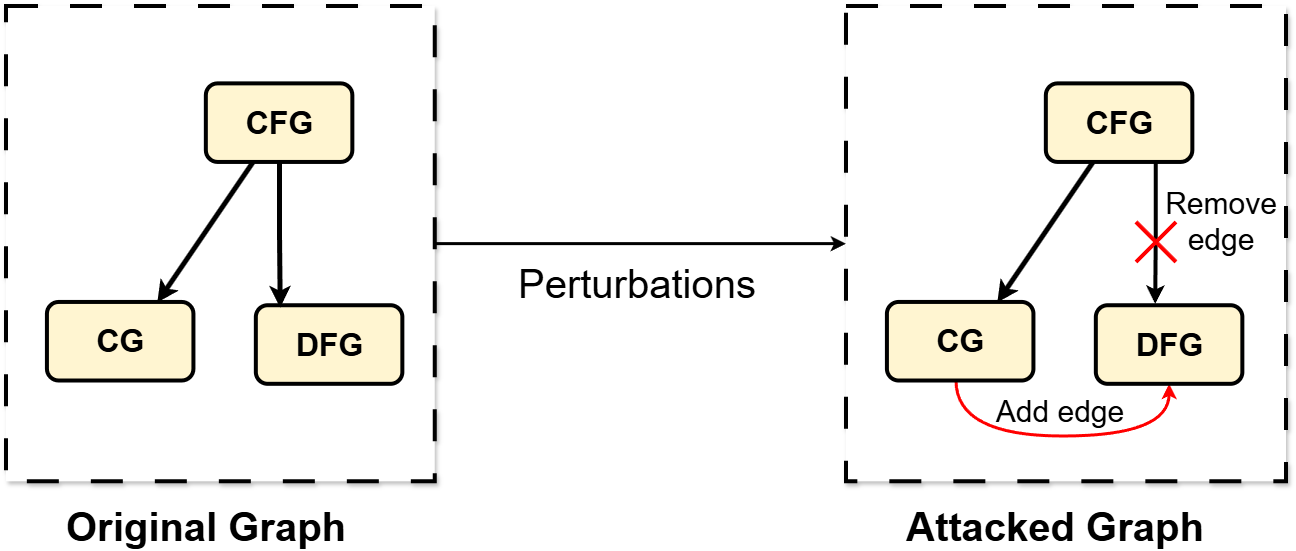}
    \caption{Structural adversarial perturbation involving edge addition and removal.}
    \label{fig:attack_graph}
\end{figure}

We then simulate two categories of adversarial attacks. Textual attacks (Figure \ref{fig:noisy_text}) perform word-level antonym substitution, replacing keywords such as \textquotedbl{}Critical\textquotedbl{} with \textquotedbl{}Safe\textquotedbl{} or \textquotedbl{}before\textquotedbl{} with \textquotedbl{}after\textquotedbl{}, preserving grammatical structure while reversing semantic meaning. Structural attacks (Figure \ref{fig:attack_graph}) involve adding spurious edges or removing critical execution paths within the control- and data-flow graph.

We select three representative hard-label black-box evasion attacks, each assuming no access to model internals:
\begin{itemize}
    \item \textbf{HSAttack \cite{li2025metapath}:} Applied to heterogeneous architectures (ORACAL, SCVHunter, MANDO-HGT). HSAttack partitions the graph into single-edge-type subgraphs and iteratively adds or removes high-impact edges to assess structural stability without requiring predefined metapaths.
    \item \textbf{CAMA \cite{wang2023revisiting}:} Applied to homogeneous models (GNN-SC, MTVHunter). CAMA generates node-level importance maps to localize and exploit structural vulnerabilities with minimal perturbation.
    \item \textbf{SubAttack \cite{hua2026subattack}:} Applied to text-based modalities (ORACAL-enrich Causal, ORACAL-node-edge Causal). SubAttack identifies semantic indicator keywords and performs antonym substitution while preserving readability.
\end{itemize}

\textbf{Detection performance comparison.} Table \ref{table:evasion_attack_results} (Original setting) summarizes the detection accuracy of ORACAL compared to SOTA methods across the three datasets. ORACAL-node-edge Causal consistently achieves the highest Macro F1 scores (90.48\% on SoliAudit, 90.83\% on CGTWeakness, and 72.82\% on DAppScan), outperforming MANDO-HGT (82.63\%), MTVHunter (78.10\%), and SCVHunter, which exhibits notably lower performance on the real-world DAppScan dataset (28.44\%). Modality-specific analysis reveals that while structural information (\textit{ORACAL-base Causal}) is effective for curated benchmarks, semantic enrichment (\textit{ORACAL-enrich Causal}) provides superior invariance on complex, out-of-distribution datasets. By fusing both modalities, \textit{ORACAL-node-edge Causal} achieves 72.82\% F1 on DAppScan, representing a +10.82\% lead over MANDO-HGT, which demonstrates that multimodal integration is essential for generalizing to industrial-grade contracts.

\begin{figure*}[t]
    \centering
    \begin{minipage}{0.33\textwidth}
        \centering
        \includegraphics[width=\linewidth]{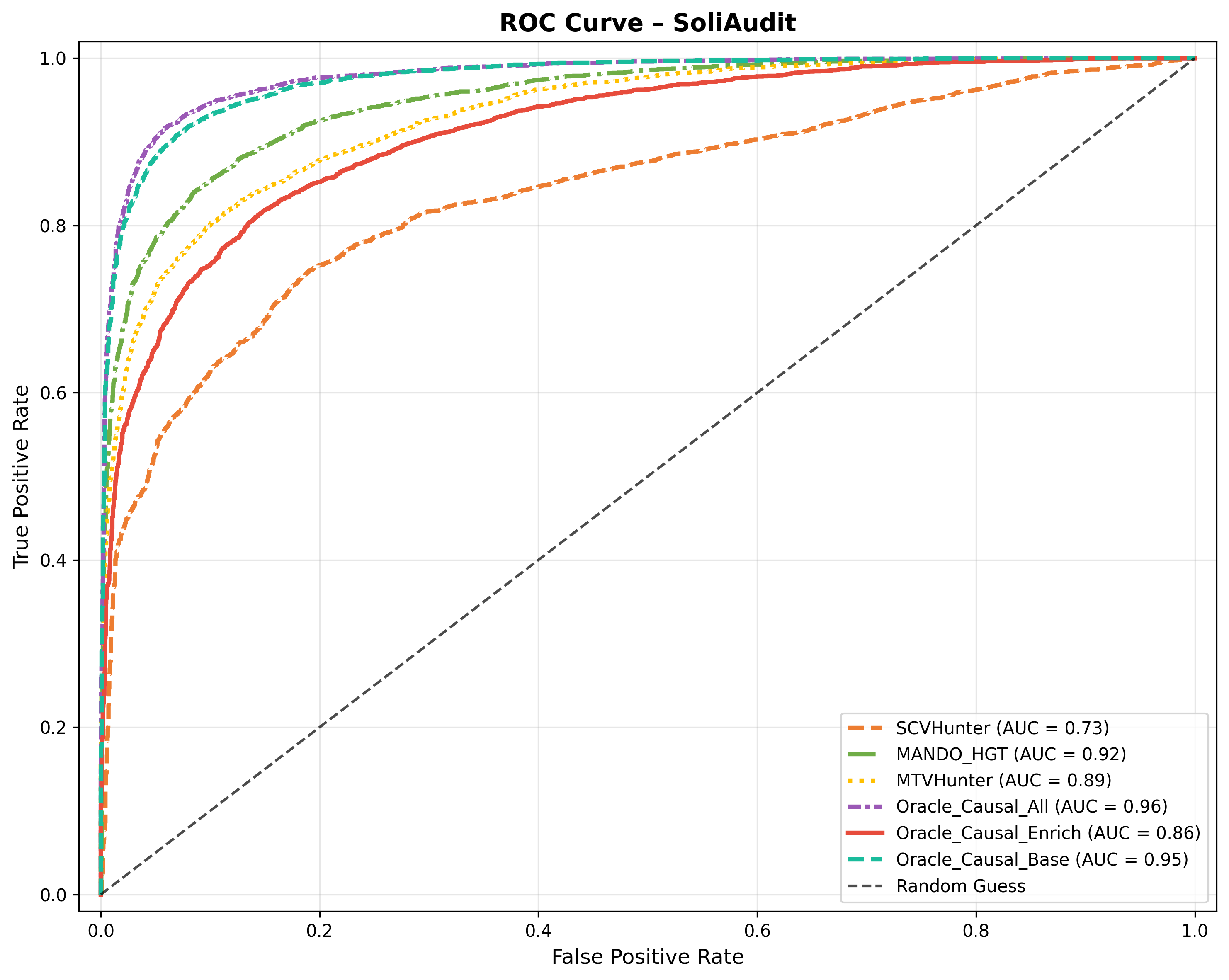}
        \caption{ROC Curve - SoliAudit}
    \end{minipage}\hfill
    \begin{minipage}{0.33\textwidth}
        \centering
        \includegraphics[width=\linewidth]{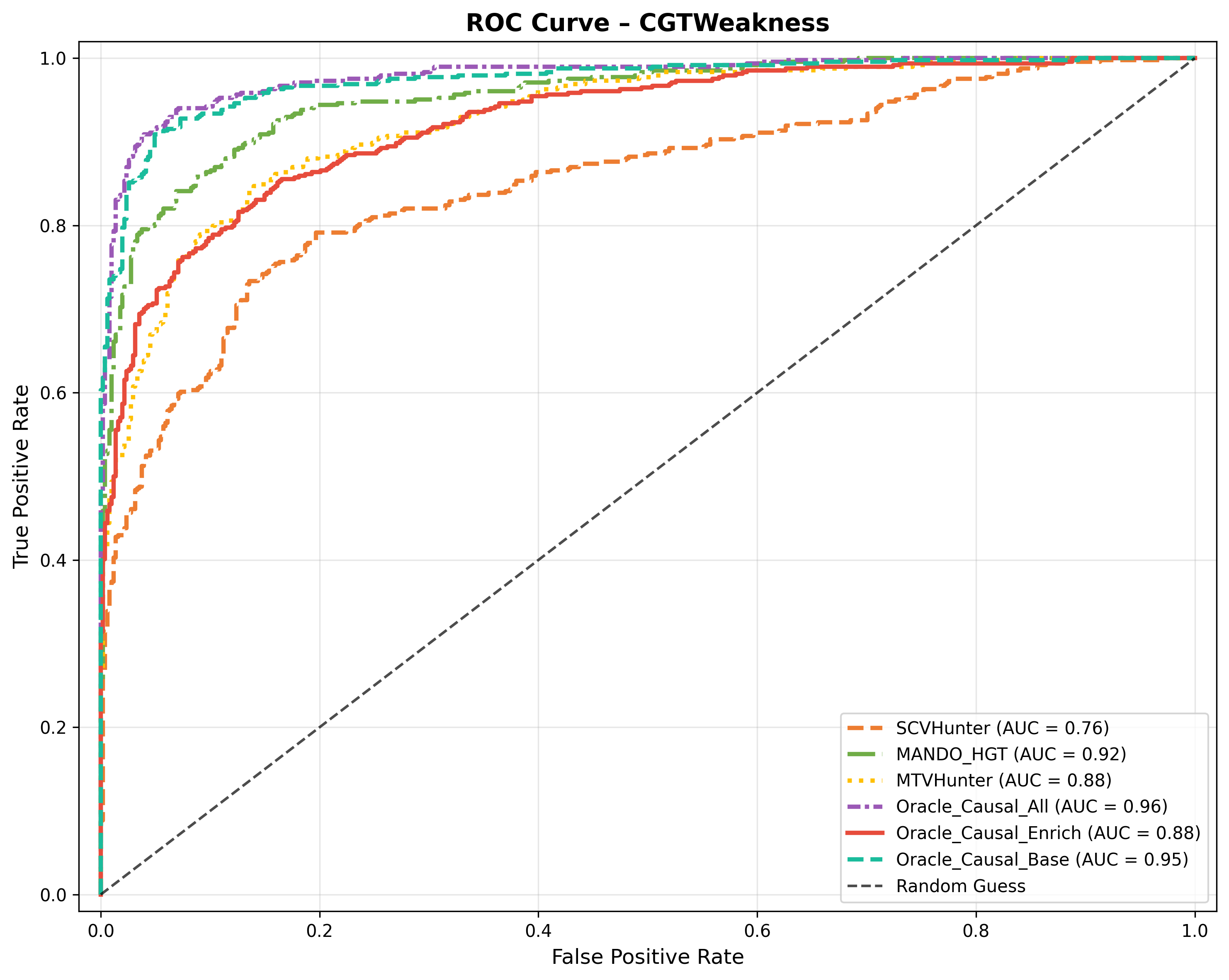}
        \caption{ROC Curve - CGTWeakness}
    \end{minipage}\hfill
    \begin{minipage}{0.33\textwidth}
        \centering
        \includegraphics[width=\linewidth]{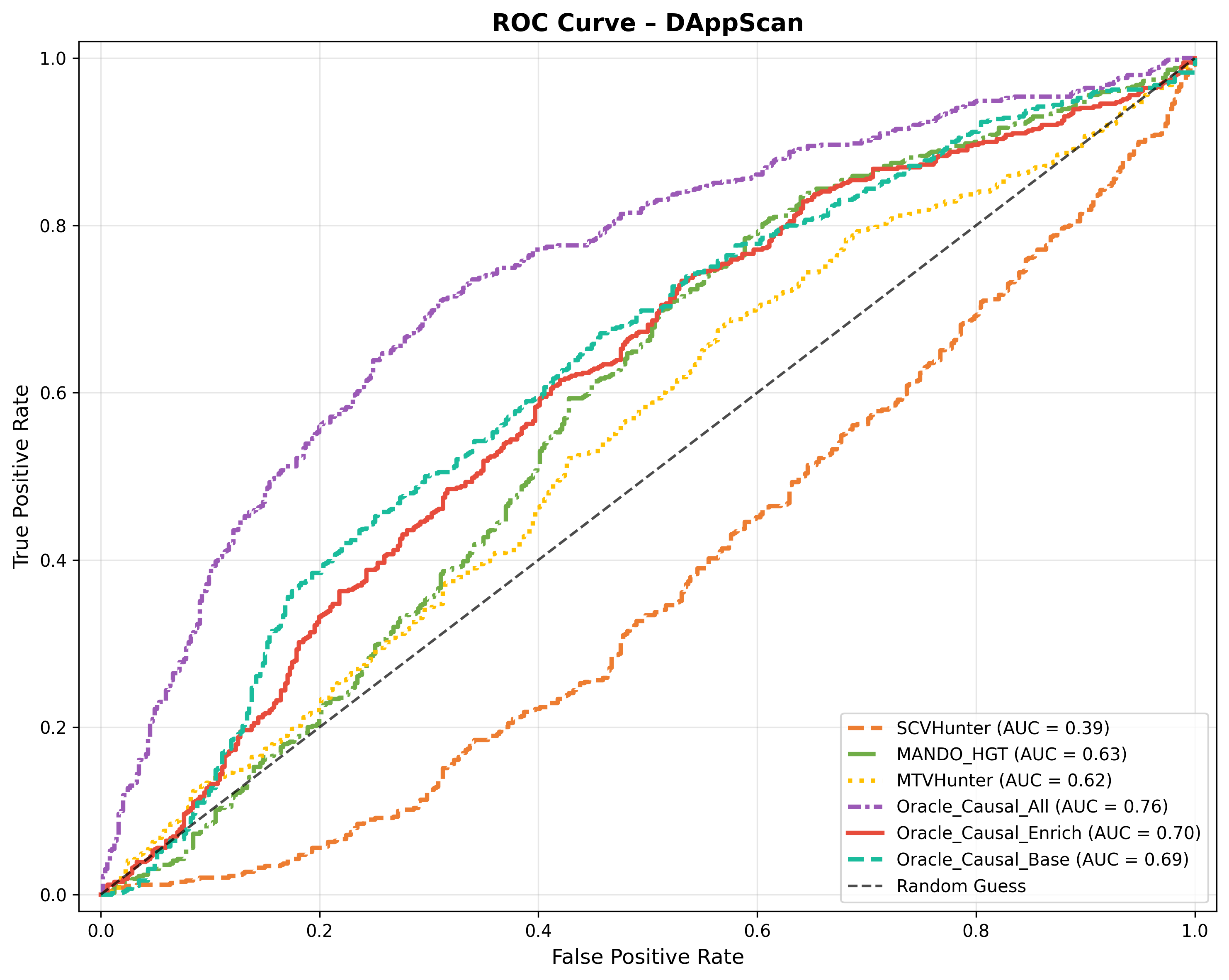}
        \caption{ROC Curve - DAppScan}
    \end{minipage}
    \caption{ROC-AUC comparison across datasets.}
    \label{fig:roc_comparison}
\end{figure*}

\textbf{ROC analysis.} Figure \ref{fig:roc_comparison} further corroborates these findings. On SoliAudit and CGTWeakness, ORACAL-node-edge Causal achieves an AUC of 0.96, while SCVHunter lags at 0.73. The performance gap is most pronounced on DAppScan, where SCVHunter's AUC drops to 0.39, whereas ORACAL-node-edge Causal maintains an AUC of 0.76. Across all datasets, ORACAL's steep initial curve near the y-axis indicates a consistently high True Positive Rate at low False Positive Rates, making it practical for security auditing tools that seek to minimize unnecessary manual review overhead.

\begin{table*}[t]
\centering
\caption{Evasion attack performance comparison. Values in parentheses indicate the relative F1 decrease vs.~the Original setting. The light-red row indicates the method achieving the highest Original F1 score (ORACAL-node-edge Causal). The dark-red row indicates the attack scenario with the lowest F1 degradation and ASR (ORACAL-node-edge Causal under SubAttack). \textquotedbl{}--\textquotedbl{} denotes not applicable (original, pre-attack setting; ASR is undefined).}
\label{table:evasion_attack_results}
\resizebox{\textwidth}{!}{
\begin{tabular}{llcccccc}
\toprule

\multirow{2}{*}{\textbf{Model}} & \multirow{2}{*}{\textbf{Method}} 
& \multicolumn{2}{c}{\textbf{SoliAudit}}
& \multicolumn{2}{c}{\textbf{CGTWeakness}}
& \multicolumn{2}{c}{\textbf{DAppScan}} \\

\cmidrule(lr){3-4}
\cmidrule(lr){5-6}
\cmidrule(lr){7-8}

& 
& \textbf{F1(\%)} & \textbf{ASR(\%)}
& \textbf{F1(\%)} & \textbf{ASR(\%)}
& \textbf{F1(\%)} & \textbf{ASR(\%)} \\

\midrule

\multirow{2}{*}{\textbf{GNN-SC}}

& Original
& 70.74 & -
& 70.20 & -
& 66.47 & - \\

& HSAttack
& 61.92(-8.82) & 11.04
& 61.07(-9.13) & 11.62
& 57.90(-8.57) & 10.31 \\

\midrule

\multirow{2}{*}{\textbf{MTVHunter}}

& Original
& 78.10 & -
& 78.73 & -
& 58.94 & - \\

& HSAttack
& 69.65(-8.45) & 10.91
& 69.82(-8.91) & 11.47
& 51.92(-7.02) & 9.73 \\

\midrule

\multirow{2}{*}{\textbf{SCVHunter}}

& Original
& 51.62 & -
& 50.20 & -
& 28.44 & - \\

& HSAttack
& 33.04(-18.58) & 18.73
& 32.71(-17.49) & 19.16
& 17.29(-11.15) & 12.02 \\

\midrule

\multirow{2}{*}{\textbf{MANDO-HGT}}

& Original
& 82.63 & -
& 83.77 & -
& 62.00 & - \\

& HSAttack
& 72.94(-9.69) & 12.41
& 73.15(-10.62) & 13.07
& 54.47(-7.53) & 10.11 \\

\midrule

\multirow{2}{*}{\textbf{ORACAL-base Causal}}

& Original
& 89.35 & -
& 90.32 & -
& 63.65 & - \\

& HSAttack
& 85.54(-3.81) & 4.22
& 86.33(-3.99) & 4.51
& 60.47(-3.18) & 4.63 \\

\midrule

\multirow{2}{*}{\textbf{ORACAL-enrich Causal}}

& Original
& 74.60 & -
& 76.83 & -
& 68.10 & - \\

& SubAttack
& 71.18(-3.42) & 4.68
& 73.02(-3.81) & 4.94
& 64.61(-3.49) & 5.03 \\

\midrule

\multirow{4}{*}{\textbf{ORACAL-node-edge Causal}}

& \cellcolor{stdred}Original
& \cellcolor{stdred}90.48 & \cellcolor{stdred}-
& \cellcolor{stdred}90.83 & \cellcolor{stdred}-
& \cellcolor{stdred}72.82 & \cellcolor{stdred}- \\

& HSAttack
& 88.13(-2.35) & 2.94
& 88.07(-2.76) & 3.21
& 70.64(-2.18) & 2.88 \\

& \cellcolor{causalred}SubAttack
& \cellcolor{causalred}88.81(-1.67) & \cellcolor{causalred}2.13
& \cellcolor{causalred}88.99(-1.84) & \cellcolor{causalred}2.27
& \cellcolor{causalred}71.71(-1.11) & \cellcolor{causalred}1.96 \\

& Both
& 85.91(-4.57) & 5.71
& 85.36(-5.47) & 6.04
& 68.61(-4.21) & 5.31 \\

\bottomrule
\end{tabular}
}
\end{table*}

\textbf{Robustness results.} ORACAL-node-edge Causal demonstrates strong resilience against all attack types. Under HSAttack, it suffers only 2.35\% F1 degradation on SoliAudit (ASR 2.94 to 3.21\%), compared to 18.58\% for SCVHunter (ASR 18.73\%). Against SubAttack, it achieves the lowest ASR of approximately 2\% and minimal F1 degradation of 1.11 to 1.84\% across all datasets. Even under simultaneous structural and textual attacks, F1 degrades by only 4.57\% on SoliAudit, a margin still smaller than any single structural attack on SOTA baselines (e.g., 9.69\% for MANDO-HGT). This robustness stems from two core mechanisms: (i) \textit{Multimodal Redundancy}, where fusing graph topology with semantic enrichment provides a fail-safe so that corrupting one modality is compensated by the other; and (ii) \textit{Causal Attention Learning}, which disentangles causal invariant features from spurious correlations, filtering out adversarial noise that does not correspond to true vulnerability causes.

\begin{tcolorbox}[colback=white,colframe=black,title=Answer to RQ3]
ORACAL-node-edge Causal achieves the highest detection accuracy (90.83\% F1 on CGTWeakness, ROC-AUC up to 0.96) and superior adversarial robustness compared to SOTA: while SCVHunter degrades by up to 18.58\% F1 under HSAttack (ASR 18.73\%), ORACAL-node-edge Causal limits degradation to 2.35\% (ASR $\sim$3\%). This resilience stems from ORACAL's multimodal redundancy and causal attention, which filter out structural and semantic adversarial noise without relying on spurious correlations.
\end{tcolorbox}
\section{Threats to Validity}
\label{sec:threats_validity}

This section discusses threats to internal, external, construct, and conclusion validity, along with our mitigation strategies.

\subsection{Internal Validity}
To address internal validity, we identify factors that could affect the accuracy of our methodology and implement corresponding controls:
\begin{itemize}
    \item \textbf{Data Preprocessing and Extraction:} The reliance on compilation and AST extraction may bias our datasets toward well-structured contracts. We mitigate this by following standard preprocessing pipelines and ensuring our extraction scripts are deterministic for reproducibility.

    \item \textbf{RAG Enrichment and LLM Stochasticity:} Semantic enrichment quality depends on corpus composition and LLM stochasticity. To control for variability, we fix all random seeds, use zero temperature for LLM inference, construct the corpus from authoritative sources, and apply a two-stage retrieval with re-ranking.

    \item \textbf{Checkpoint Selection for Generalization:} Evaluating all configurations is computationally expensive, so generalization evaluation uses only the top-performing checkpoint from each paradigm. We mitigate this limitation by selecting checkpoints based on primary benchmark performance to represent practically relevant models.
\end{itemize}

\subsection{External Validity}
Regarding external validity, we consider the extent to which our findings can be generalized to different contexts and future scenarios:
\begin{itemize}
    \item \textbf{Generalizability:} Our evaluation focuses on Solidity datasets covering Decentralized Application Security Project (DASP) Top 10 taxonomy, which may not transfer to other languages or novel vulnerabilities ranging from flash loans to MEV. We address this by evaluating across structurally diverse datasets, including two independent out-of-distribution datasets (DAppScan and LLMAV), demonstrating generalization across distinct collection and labeling methodologies.

    \item \textbf{Temporal Concept Drift:} The model is trained on static snapshots, risking performance degradation as the Solidity ecosystem evolves and new vulnerabilities emerge. We partially mitigate this through ORACAL's modular RAG corpus, which can be updated independently, and its causal attention mechanism designed to learn invariant semantics, improving resilience to distributional shifts.
\end{itemize}

\subsection{Construct Validity}
Construct validity concerns whether our chosen measurements and labels accurately reflect the theoretical concepts being studied. We address the following challenges:
\begin{itemize}
    \item \textbf{Label Reliability:} Ground truth labels are derived from heterogeneous sources, introducing potential noise. We mitigate this by using consolidated datasets with multi-tool voting and cross-referencing automated labels with manual case studies to validate semantic consistency.

    \item \textbf{Explainability Ground Truth:} Explainability evaluation relies on LLMAV's manual line-level annotations, involving subjective boundary decisions. We address this by evaluating on both LLMAV and DAppScan, demonstrating consistent explainer rankings across independent datasets, reducing the likelihood of methodology artifacts.

    \item \textbf{Adversarial Budget Scope:} We evaluate robustness under fixed perturbation budgets, while real-world adversaries may operate under different constraints. We acknowledge this boundary condition and note that evaluating a range of budgets would provide a more complete robustness profile.

    \item \textbf{Evaluation Metrics:} Standard metrics may not capture all aspects of practical auditing (specifically, Macro F1 treats all classes equally regardless of severity). We supplement aggregate metrics with per-class breakdowns, qualitative case studies, and structural bias analyses.
\end{itemize}

\subsection{Conclusion Validity}
Conclusion validity focuses on the statistical and analytical rigor of our inferences. We account for the following potential issues:
\begin{itemize}
    \item \textbf{Statistical Soundness:} The out-of-distribution test sets are relatively small, and Wilcoxon signed-rank tests ($n=5$) yield $p$-values constrained by a discrete distribution. We complement $p$-values with Vargha-Delaney $\hat{A}_{12}$ effect sizes, where consistently maximal values ($\hat{A}_{12} = 1.00$) provide strong evidence of systematic gains.

    \item \textbf{Implicit Overfitting through Iterative Design:} Repeated evaluation may inadvertently guide architectural choices toward test-specific optimizations. We mitigate this by maintaining strict train-test separation, reserving two independent OOD datasets (DAppScan and LLMAV) unused during model selection, and reporting performance across multiple independent benchmarks.
\end{itemize}

\section{Conclusion and Future Work}
\label{sec:conclusion}

In this paper, we presented ORACAL, a novel framework that bridges graph-based structural analysis and LLM-based semantic reasoning for smart contract vulnerability detection. By enriching a heterogeneous graph with \textquotedbl{}Operational Context\textquotedbl{} and \textquotedbl{}Security Analysis\textquotedbl{} via a trusted RAG pipeline, ORACAL achieves a Macro F1 of 91.28\% on SoliAudit and consistently outperforms the standard paradigm by between 3.5 and 6.2 percentage points across CGT, DAppScan, and LLMAV benchmarks, driven by a Causal Attention mechanism that disentangles true vulnerability signals from spurious correlations. Beyond detection, ORACAL delivers auditor-friendly explanations via PGExplainer (up to 32.51\% MIoU on LLMAV) and strong adversarial robustness, maintaining an Attack Success Rate of only 2 to 3\% under both structural and textual perturbations, compared to up to 18.73\% for existing methods such as SCVHunter.

Future work will focus on three directions: (1) extending the framework to Node Classification to pinpoint the exact locations of vulnerabilities within the code (line-level detection), (2) employing established evaluation metrics \cite{yu2024evaluation} to systematically assess the faithfulness and relevance of RAG-generated semantic descriptions, ensuring that the enriched content remains grounded in technical truth, and (3) exploring Incremental Learning techniques \cite{leo2024survey} to efficiently integrate new samples without full retraining. ORACAL represents a significant step towards transparent, automated, and explainable security auditing for the blockchain ecosystem.












\bibliographystyle{unsrt}

\bibliography{refs.bib}

\end{document}